%% file: main.tex
\documentclass[lettersize,journal]{IEEEtran}
\usepackage{amsmath,amsfonts}
\usepackage{algorithmic}
\usepackage{array}
\usepackage[caption=false,font=normalsize,labelfont=sf,textfont=sf]{subfig}
\usepackage{textcomp}
\usepackage{stfloats}
\usepackage{url}
\usepackage{verbatim}
\usepackage{graphicx}
\usepackage{cite}
\usepackage[backref=page]{hyperref}
\usepackage{cleveref}
\usepackage{booktabs}
\usepackage{tikz}
\usepackage[edges]{forest}
\usetikzlibrary{mindmap}
\usetikzlibrary{trees}
\definecolor{hidden-draw}{RGB}{20,68,106}
\definecolor{hidden-pink}{RGB}{255,245,247}
\def\BibTeX{{\rm B\kern-.05em{\sc i\kern-.025em b}\kern-.08em
    T\kern-.1667em\lower.7ex\hbox{E}\kern-.125emX}}
\usepackage{balance}
\begin{document}
\title{Visual Large Language Models for  Generalized and Specialized Applications}

\author{Yifan Li\textsuperscript{1}, Zhixin Lai\textsuperscript{2}, Wentao Bao\textsuperscript{1}, Zhen Tan\textsuperscript{3}, Anh Dao\textsuperscript{1}, Kewei Sui\textsuperscript{4}, \\Jiayi Shen\textsuperscript{5}, Dong Liu\textsuperscript{6}, Huan Liu\textsuperscript{3}, Yu Kong\textsuperscript{1} \\
\textsuperscript{1}Michigan State University, \texttt{\{liyifa11, baowenta, anhdao, yukong\}@msu.edu} \\
\textsuperscript{2}Cornell University, \texttt{zl768@cornell.edu} \\
\textsuperscript{3}Arizona State University, \texttt{\{ztan36, huanliu\}@asu.edu} \\
\textsuperscript{4}University of California, Berkeley, \texttt{ksui2@berkeley.edu} \\
\textsuperscript{5}University of Texas at Austin, \texttt{sjyiya@gmail.com} \\
\textsuperscript{6}Yale University, \texttt{pikeliu.misys@gmail.com}

}

\markboth{Journal of \LaTeX\ Class Files, January~2025}%
{How to Use the IEEEtran \LaTeX \ Templates}

\maketitle

\input{contents/abstract}
\input{contents/intro}
\section{Texonomy}
\input{figures/taxonomy}
\input{contents/image2text}
\input{contents/video2text}
\input{contents/vision2action}

\input{contents/text2vision}
\input{tables/datasets}
\input{contents/challenges}

\input{contents/future_direction}

\bibliographystyle{IEEEtran}
\bibliography{main,yifan,anh,wentao,zhen,zhixin,kewei,dong,jiayi}
\end{document}

%% file: contents/abstract.tex
\begin{abstract}

Visual-language models (VLM) have emerged as a powerful tool for learning a unified embedding space for vision and language. Inspired by large language models, which have demonstrated strong reasoning and multi-task capabilities, visual large language models (VLLMs) are gaining increasing attention for building general-purpose VLMs. Despite the significant progress made in VLLMs, the related literature remains limited, particularly from a {\textit{comprehensive application}} perspective, encompassing generalized and specialized applications across vision (image, video, depth), action, and language modalities. In this survey, we focus on the diverse applications of VLLMs, examining their using scenarios, identifying ethics consideration and challenges, and discussing future directions for their development. By synthesizing these contents, we aim to provide a comprehensive guide that will pave the way for future innovations and broader applications of VLLMs. The paper list repository is available: \url{https://github.com/JackYFL/awesome-VLLMs}.
\end{abstract}

\begin{IEEEkeywords}
Vision large language model (VLLM), Large Language Model (LLM), vision language model (VLM), applications, multi-modal learning
\end{IEEEkeywords}

%% file: contents/intro.tex
\section{Introduction}

Computer vision tasks are challenging and diverse, requiring machines to possess a range of capabilities such as object perception \cite{wang2017fast, he2016deep}, spatial  understanding \cite{huang2023voxposer}, temporal action interpretation \cite{feichtenhofer2017spatiotemporal, ding2023temporal}, interaction with humans \cite{antol2015vqa}, and the ability to handle various domains and data transformations, among others. Numerous vision-language models (VLMs) \cite{ding2023temporal, zhang2024vision} have been developed to enhance the generalization and reasoning abilities of vision models by leveraging large-scale vision-language data, thereby improving their capacity to address these diverse challenges. 
\begin{figure}
    \centering
    \includegraphics[width=1\linewidth]{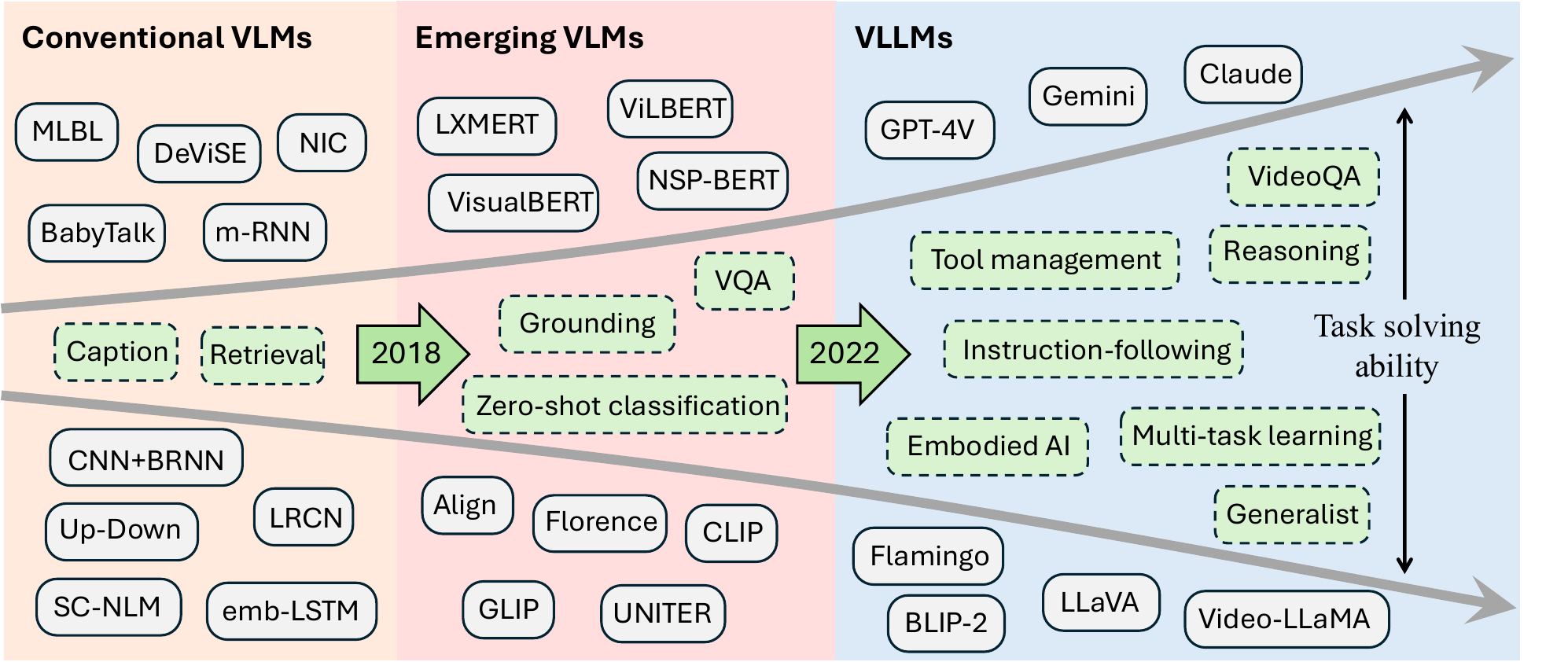}
    \caption{The evolution of VLMs includes three phases: conventional VLMs (before 2018), emerging VLMs (2018-2022), and VLLMs (2022 onward). }
    \label{fig:vlm_trends}
\end{figure}

Conventional VLMs (before 2018, \cref{fig:vlm_trends}) are designed to handle limited vision tasks, such as captioning and retrieval, primarily through encoder-decoder architectures. These approaches rely on convolutional neural networks (CNNs) to extract visual representations \cite{fang2015captions, vinyals2015show} or hand-crafted visual features,  and generate sentences using recurrent neural networks (RNNs) \cite{karpathy2015deep}, long short-term memory (LSTM) \cite{jia2015guiding} or conditional random field (CRF) \cite{kulkarni2013babytalk}-based decoder. However, constrained by the limited scale of pretrained datasets and the simplicity of their architectures, these models are only capable of addressing specific tasks with moderate performance.

With the advancement of deep learning techniques such as Transformer \cite{Vaswani2017AttentionIA}, research on Vision-Language Models (VLMs) gained emergence from 2018 to 2022 (see \cref{fig:vlm_trends}).  
Inspired by breakthroughs in natural language processing, these VLMs typically follow the \textit{Pre-training, Fine-tuning, and Prediction} paradigm across various tasks, leveraging foundation models like BERT \cite{Devlin2019BERTPO} and CLIP \cite{radford2021learning} that are pretrained on large-scale language or vision-language datasets. These foundation models encapsulate rich prior knowledge, enabling robust zero-shot and transfer learning performance across diverse tasks. Based on language foundation models like BERT, numerous approaches \cite{li2019visualbert,lu2019vilbert,sun2022nsp} have been developed by integrating a visual extractor with the language foundation model, replacing earlier recurrent architectures. By leveraging the prior knowledge embedded in large-scale language corpora, these methods achieve significantly better performance compared to conventional models. Building on VLM foundation models such as CLIP, many studies aim to transfer this prior knowledge to downstream tasks through techniques like prompt tuning \cite{zhou2022learning, zhou2022conditional}, visual adaptation \cite{mu2022slip}, and knowledge distillation \cite{ding2022decoupling, guopen}. Despite these advancements, most of these methods are discriminative-based, which inherently lack multi-task capabilities and strong reasoning abilities. This limitation hinders the further development and broader applications of VLMs.

\begin{figure}
    \centering
    \includegraphics[width=1\linewidth]{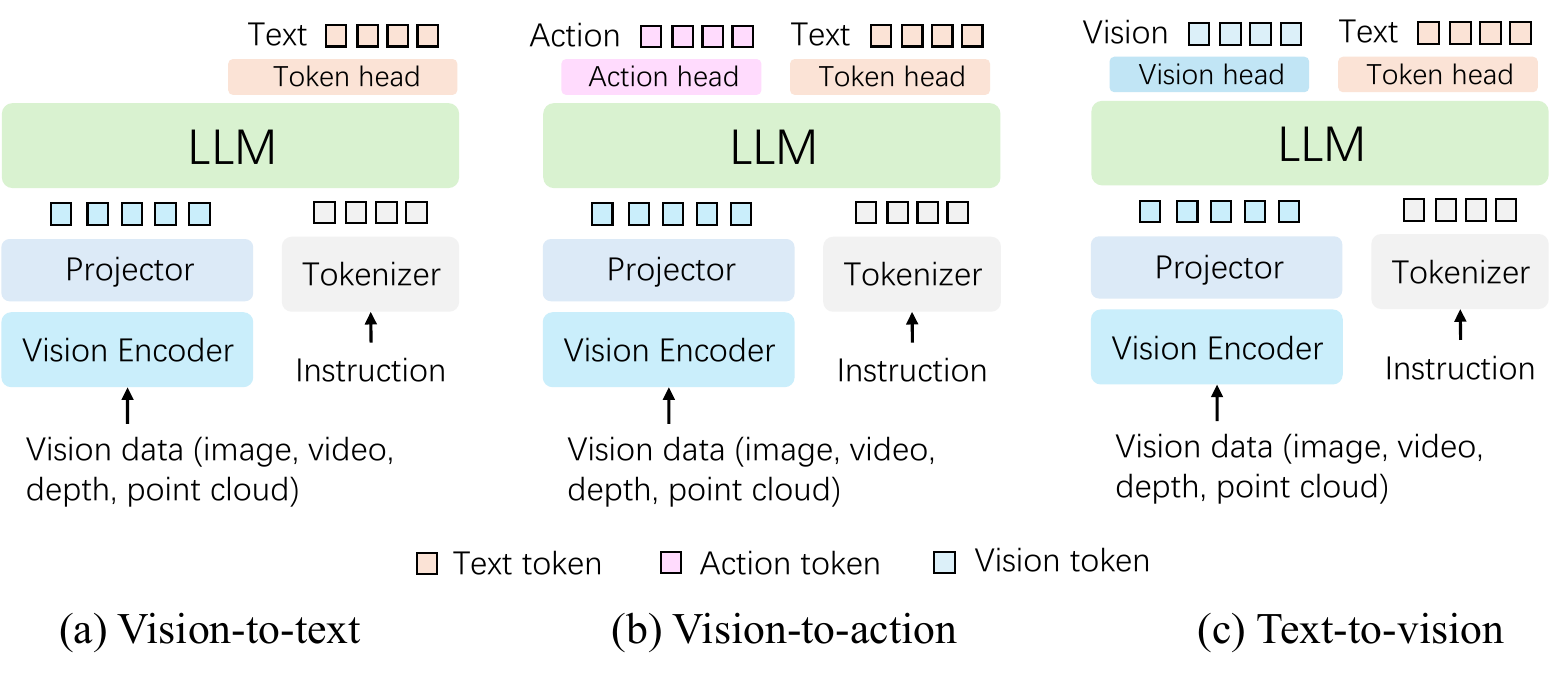}
    \caption{The architectures of three VLLM application categories: (a) vision-to-text, (b) vision-to-action, (c) text-to-vision.}
    \label{fig:category}
\end{figure}

With the rise of language generative models \cite{brown2020language,ouyang2022training}, researchers focus  on leveraging the prior knowledge embedded in Large Language Models (LLMs) to develop general-purpose and highly reasoning VLMs. Leveraging instruction-tuning techniques \cite{zhang2023instruction} in LLMs, current Visual Large Language Models (VLLMs) \cite{alayrac2022flamingo} can process versatile instructions and generate responses that align with human preferences. Specifically, these VLLMs (as shown in \cref{fig:category}) employ a vision encoder to patchy vision data, use a connector to project visual tokens into the language space, and rely on an LLM as a decoder to produce instruction-following answers. These VLLMs can be applied to generalized and specialized applications, leveraging the prior knowledge and the sequence-to-sequence architecture of LLMs. Furthermore, they also inherit the reasoning capabilities of LLMs, enabling them to tackle more complex vision tasks.

Previous related surveys have primarily focused on general VLMs~\cite{long2022vision, zhou2023vision, zhang2024vision, ghosh2024exploring}, or specific aspects of multi-modal large language models (MLLMs), such as  methodologies~\cite{zhang2024mm, huang2023visual}, reasoning capabilities~\cite{wang2024exploring}, interpretability~\cite{dang2024explainable}, evaluation benchmarks~\cite{fu2024mme}, development trends~\cite{yin2024survey, wu2023multimodal, caffagni2024r, li2024multimodal}, and hallucination issues~\cite{liu2024survey}. With the advancement of VLLMs across various tasks, their applications have become increasingly diverse, covering both generalized and specialized scenarios. However, these surveys lack a comprehensive overview of the applications, limitations, and future directions of the vision modality in MLLMs. 

In contrast, our survey provides a holistic perspective on VLLM applications, categorizing them into three main aspects: \textit{vision-to-text}, \textit{vision-to-action}, and \textit{text-to-vision}. Each category is further broken down into detailed subtasks, with an in-depth analysis. Additionally, we explore the social impact of VLLMs, and the challenges they face in different scenarios. Finally, based on the ethical concerns, we propose some promising directions for future development, like security \& privacy, efficiency, interpretability \& explaninability, complex reasoning, \textit{etc}. We hope our survey will provide a fresh perspective on VLLM applications and serve as a continuous source of inspiration for others.

\vspace{-10pt}






%% file: figures/taxonomy.tex
\tikzstyle{my-box}=[
    rectangle,
    draw=hidden-draw,
    rounded corners,
    text opacity=1,
    minimum height=1.5em,
    minimum width=5em,
    inner sep=2pt,
    align=center,
    fill opacity=.5,
    line width=0.8pt,
]
\tikzstyle{leaf}=[my-box, minimum height=1.5em,
    fill=green!6, text=black, align=left,font=\normalsize,
    inner xsep=2pt,
    inner ysep=4pt,
    line width=0.8pt,
]

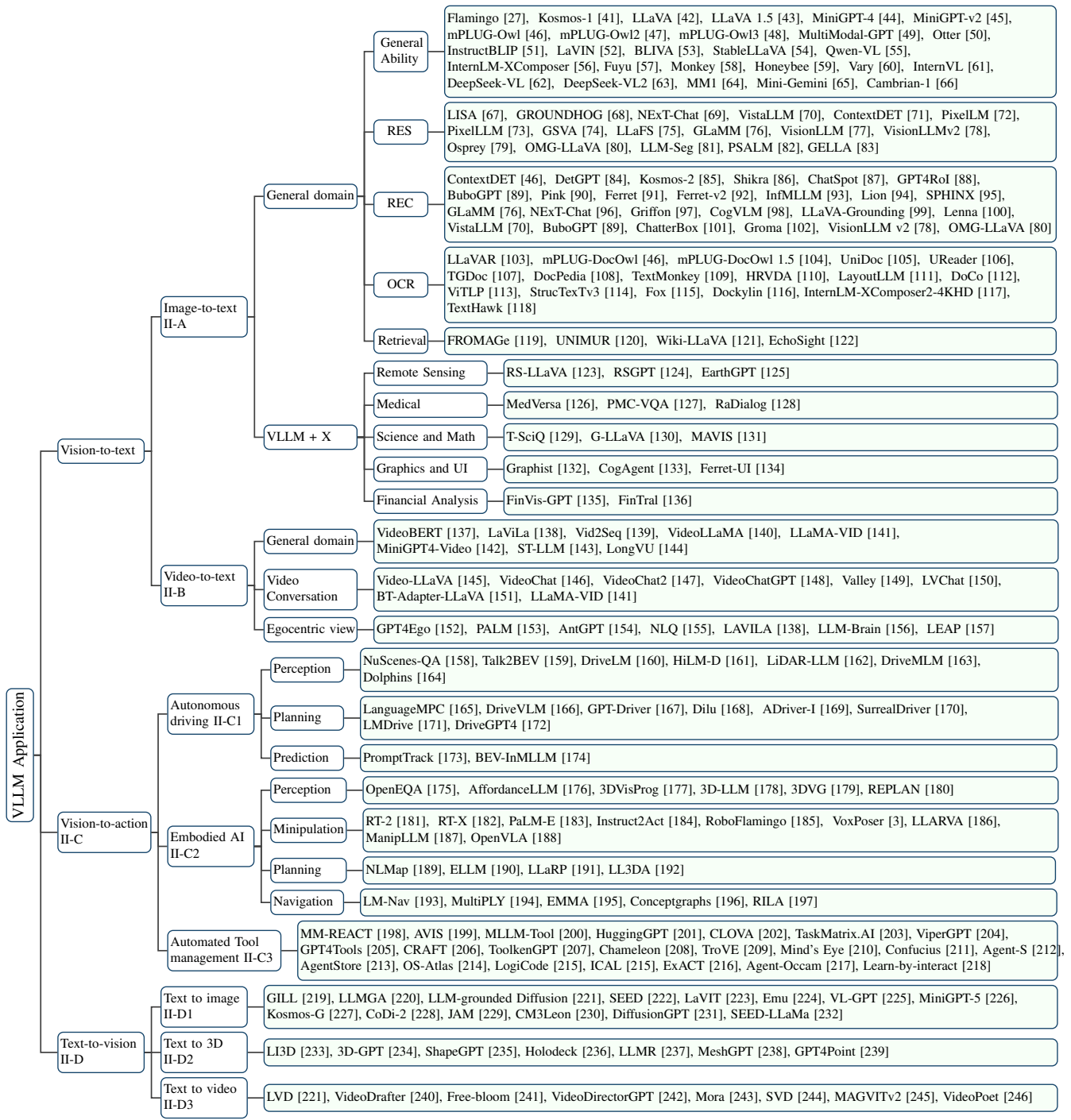
\begin{figure*}[!th]
    \centering
    \resizebox{0.98\textwidth}{!}{
        \begin{forest}
            forked edges,
            for tree={
                grow=east,
                reversed=true,
                anchor=base west,
                parent anchor=east,
                child anchor=west,
                base=left,
                font=\large,
                rectangle,
                draw=hidden-draw,
                rounded corners,
                align=left,
                minimum width=4em,
                edge+={darkgray, line width=1pt},
                s sep=3pt,
                inner xsep=2pt,
                inner ysep=3pt,
                line width=0.8pt,
                ver/.style={rotate=90, child anchor=north, parent anchor=south, anchor=center},
            },
            where level=1{text width=6em,font=\normalsize,}{},
            where level=2{text width=6em,font=\normalsize,}{},
            where level=3{text width=6.5em,font=\normalsize,}{},
            where level=4{text width=5em,font=\normalsize,}{},
            [
                VLLM Application, ver
                [
                    Vision-to-text 
                    [
                        Image-to-text \\
                        \ref{img2txt}
                        [
                            General domain
                            [
                                General \\
                                Ability, text width=3em
                                    [
                                        Flamingo~\cite{alayrac2022flamingo}{,
                                        }
                                        Kosmos-1~\cite{huang2023language}{, 
                                        }
                                        LLaVA~\cite{liu2024visual}{, }
                                        LLaVA 1.5~\cite{liu2024improved}{,
                                        }
                                        MiniGPT-4~\cite{zhu2023minigpt}{, 
                                        }
                                        MiniGPT-v2~\cite{chen2023minigpt}{, 
                                        }\\
                                        mPLUG-Owl~\cite{ye2023mplug}{,
                                        } 
                                        mPLUG-Owl2~\cite{ye2024mplug}{, 
                                        }
                                        mPLUG-Owl3~\cite{ye2024mplug3}{, 
                                        }
                                        MultiModal-GPT~\cite{gong2023multimodal}{,
                                        } 
                                        Otter~\cite{li2023otter}{,
                                        } \\
                                        InstructBLIP~\cite{dai2023instructblip}{,
                                        }
                                        LaVIN~\cite{luo2024cheap}{,
                                        } 
                                        BLIVA~\cite{hu2024bliva}{,
                                        }
                                        StableLLaVA~\cite{li2023stablellava}{,
                                        }
                                        Qwen-VL~\cite{bai2023qwen}{,
                                        } \\
                                        InternLM-XComposer~\cite{zhang2023internlm}{,} 
                                        Fuyu~\cite{fuyu-8b}{,
                                        }
                                        Monkey~\cite{li2024monkey}{, }
                                        Honeybee~\cite{cha2024honeybee}{,
                                        } 
                                        Vary~\cite{wei2023vary}{,
                                        } 
                                        InternVL~\cite{chen2024internvl}{,
                                        } \\
                                        DeepSeek-VL~\cite{lu2024deepseek}{,
                                        } 
                                        DeepSeek-VL2~\cite{wu2024deepseek}{,
                                        }
                                        MM1~\cite{mckinzie2024mm1}{,
                                        } 
                                        Mini-Gemini~\cite{li2024mini}{,
                                        }
                                        Cambrian-1~\cite{tong2024cambrian}
                                        ,
                                        leaf, text width=45em
                                    ]
                            ]
                            [
                                RES, text width=2em
                                    [
                                        LISA~\cite{lai2024lisa}{,
                                        }
                                        GROUNDHOG~\cite{zhang2024groundhog}{, }NExT-Chat~\cite{zhang2023next}{, 
                                        }
                                        VistaLLM~\cite{pramanick2024jack}{,
                                        } 
                                        ContextDET~\cite{zang2023contextual}{,
                                        } 
                                        PixelLM~\cite{ren2024pixellm}{,
                                        } \\
                                        PixelLLM~\cite{xu2024pixel}{,
                                        }
                                        GSVA~\cite{xia2024gsva}{,
                                        } 
                                        LLaFS~\cite{zhu2024llafs}{,
                                        }
                                        GLaMM~\cite{rasheed2024glamm}{,
                                        } 
                                        VisionLLM~\cite{wang2024visionllm}{,
                                        } 
                                        VisionLLMv2~\cite{wu2024visionllm}{,
                                        } \\ 
                                        Osprey~\cite{yuan2024osprey}{,
                                        }
                                        OMG-LLaVA~\cite{zhang2024omg}{,
                                        }
                                        LLM-Seg~\cite{wang2024llm}{,}
                                        PSALM~\cite{zhang2024psalm}{,}
                                        GELLA~\cite{qi2024generalizable},
                                        leaf, text width=45em
                                    ]
                            ]
                            [
                                REC, text width=2em
                                    [
                                        ContextDET~\cite{ye2023mplug}{, 
                                        }
                                        DetGPT~\cite{pi2023detgpt}{, }
                                        Kosmos-2~\cite{peng2023kosmos}{, 
                                        }
                                        Shikra~\cite{chen2023shikra}{,
                                        } 
                                        ChatSpot~\cite{zhao2023chatspot}{,
                                        } 
                                        GPT4RoI~\cite{zhang2023gpt4roi}{,
                                        } \\
                                        BuboGPT~\cite{zhao2023bubogpt}{,
                                        } 
                                        Pink~\cite{xuan2024pink}{,
                                        } 
                                        Ferret~\cite{you2023ferret}{,
                                        } 
                                        Ferret-v2~\cite{zhang2024ferret}{,
                                        } 
                                        InfMLLM~\cite{zhou2023infmllm}{,
                                        }
                                        Lion~\cite{chen2024lion}{,
                                        } 
                                        SPHINX~\cite{lin2023sphinx}{,
                                        } \\
                                        GLaMM~\cite{rasheed2024glamm}{,} 
                                        NExT-Chat~\cite{head2next}{,
                                        } 
                                        Griffon~\cite{zhan2023griffon}{,
                                        } 
                                        CogVLM~\cite{wang2023cogvlm}{, 
                                        } 
                                        LLaVA-Grounding~\cite{zhang2023llava}{, }
                                        Lenna~\cite{wei2023lenna}{, 
                                        } \\
                                        VistaLLM~\cite{pramanick2024jack}{,
                                        } 
                                        BuboGPT~\cite{zhao2023bubogpt}{,
                                        } 
                                        ChatterBox~\cite{tian2024chatterbox}{,
                                        } 
                                        Groma~\cite{ma2024groma}{,
                                        } 
                                        VisionLLM v2~\cite{wu2024visionllm}{,
                                        } 
                                        OMG-LLaVA~\cite{zhang2024omg},
                                        leaf, text width=45em
                                    ]
                            ]
                            [
                                OCR, text width=2em
                                    [
                                        LLaVAR~\cite{zhang2023llavar}{,
                                        }
                                        mPLUG-DocOwl~\cite{ye2023mplug}{, 
                                        }
                                        mPLUG-DocOwl 1.5~\cite{hu2024mplug}{,
                                        }
                                        UniDoc~\cite{fengunidoc}{, }
                                        UReader~\cite{ye2023ureader}{, 
                                        } \\
                                        TGDoc~\cite{wang2023towards}{,
                                        }
                                        DocPedia~\cite{feng2023docpedia}{,
                                        } 
                                        TextMonkey~\cite{liu2024textmonkey}{,
                                        }
                                        HRVDA~\cite{liu2024hrvda}{,
                                        } 
                                        LayoutLLM~\cite{luo2024layoutllm}{,
                                        } 
                                        DoCo~\cite{li2024enhancing}{,
                                        } \\
                                        ViTLP~\cite{mao2024visually}{,
                                        }
                                        StrucTexTv3~\cite{lyu2024structextv3}{,
                                        }
                                        Fox~\cite{liu2024focus}{,
                                        }
                                        Dockylin~\cite{zhang2024dockylin}{,}
                                        InternLM-XComposer2-4KHD~\cite{dong2024internlm}{,
                                        } \\
                                        TextHawk~\cite{yu2024texthawk},
                                        leaf, text width=45em
                                    ]
                            ]
                            [
                                Retrieval, text width=3.4em
                                    [
                                        FROMAGe~\cite{koh2023grounding}{,
                                        }
                                        UNIMUR~\cite{wang2024unified}{,
                                        }
                                        Wiki-LLaVA~\cite{caffagni2024wiki}{,}
                                        EchoSight~\cite{yan2024echosight},
                                        leaf, text width=45em
                                    ]
                            ]
                        ]
                        [
                            VLLM + X
                                [
                                    Remote Sensing, text width=8em
                                        [
                                            RS-LLaVA~\cite{rs16091477}{,
                                            }
                                            RSGPT~\cite{hu2023rsgpt}{,
                                            }
                                            EarthGPT~\cite{zhang2024earthgpt}
                                            ,
                                            leaf, text width=40.7em
                                        ]
                                ]
                                [
                                    Medical, text width=8em
                                        [
                                            MedVersa~\cite{zhou2024generalist}{,
                                            }
                                            PMC-VQA~\cite{zhang2023pmcvqa}{,
                                            }
                                            RaDialog~\cite{pellegrini2023radialog}
                                            ,
                                            leaf, text width=40.7em
                                        ]
                                ]
                                [
                                    Science and Math, text width=8em
                                        [
                                            T-SciQ~\cite{wang2023tsciq}{,
                                            }
                                            G-LLaVA~\cite{gao2023gllava}{,
                                            }
                                            MAVIS~\cite{zhang2024mavis}
                                            ,
                                            leaf, text width=40.7em
                                        ]
                                ]
                                [
                                    Graphics and UI, text width=8em
                                        [
                                            Graphist~\cite{cheng2024graphic}{,
                                            }
                                            CogAgent~\cite{hong2023cogagent}{,
                                            }
                                            Ferret-UI~\cite{you2024ferretui}
                                            ,
                                            leaf, text width=40.7em
                                        ]
                                ]
                                [
                                    Financial Analysis, text width=8em
                                        [
                                            FinVis-GPT~\cite{wang2023finvisgpt}{,
                                            }
                                            FinTral~\cite{bhatia2024fintral}
                                            ,
                                            leaf, text width=40.7em
                                        ]
                                ]
                        ]
                    ]
                    [                            
                        Video-to-text \\
                        \ref{video2text}
                        [
                            General domain
                            [
                                VideoBERT~\cite{videobert}{,
                                }
                                LaViLa~\cite{lavila}{,
                                }
                                Vid2Seq~\cite{vid2seq}{,
                                }
                                VideoLLaMA~\cite{zhang2023video}{,
                                }
                                LLaMA-VID~\cite{li2023llama}{,
                                }\\
                                MiniGPT4-Video~\cite{ataallah2024minigpt4}{,
                                }
                                ST-LLM~\cite{liu2024st}{,}
                                LongVU~\cite{shen2024longvu},
                                leaf, text width=50.4em
                            ]
                        ]
                        [
                            Video \\Conversation
                            [
                                Video-LLaVA~\cite{lin2023video}{,
                                }
                                VideoChat~\cite{li2023videochat}{,
                                }
                                VideoChat2~\cite{li2023mvbench}{,
                                }
                                VideoChatGPT~\cite{maaz2023video}{,
                                }
                                Valley~\cite{luo2023valley}{,
                                }
                                LVChat~\cite{wang2024lvchat}{, }\\
                                BT-Adapter-LLaVA~\cite{liu2023one}{,
                                }
                                LLaMA-VID~\cite{li2023llama},
                                leaf, text width=50.4em
                            ]
                        ]
                        [
                            Egocentric view
                            [
                                GPT4Ego~\cite{dai2024gpt4ego}{,
                                }
                                PALM~\cite{kim2025palm}{,
                                }
                                AntGPT~\cite{zhao2023antgpt}{,
                                }
                                NLQ~\cite{wang2023egocentric}{,
                                }
                                LAVILA~\cite{lavila}{,
                                }
                                LLM-Brain~\cite{mai2023llm}{,
                                }
                                LEAP~\cite{dessalene2023leap},
                                leaf, text width=50.4em
                            ]
                        ]
                    ]
                ]
                [
                    Vision-to-action\\
                    \ref{vision2action}, text width=6.5em
                    [
                        Autonomous \\
                        driving \ref{autonomousdriving}, text width=6em
                        [
                            Perception, text width=5em
                            [
                                NuScenes-QA~\cite{qian2024nuscenes}{, }Talk2BEV~\cite{dewangan2023talk2bev}{, }DriveLM~\cite{sima2023drivelm}{, }HiLM-D~\cite{ding2023hilm}{, }
                                LiDAR-LLM~\cite{yang2023lidar}{, }DriveMLM~\cite{wang2023drivemlm}{, }\\Dolphins~\cite{ma2023dolphins}, leaf, text width=51.4em
                            ]
                        ]
                        [                        
                            Planning, text width=5em
                            [
                                LanguageMPC~\cite{sha2023languagempc}{, }DriveVLM~\cite{tian2024drivevlm}{, }GPT-Driver~\cite{mao2023gpt}{, }Dilu~\cite{wen2023dilu}{, }
                                ADriver-I~\cite{jia2023adriver}{, }SurrealDriver~\cite{jin2023surrealdriver}{, }\\LMDrive~\cite{shao2023lmdrive}{, }DriveGPT4~\cite{xu2023drivegpt4}, leaf, text width=51.4em
                            ]
                        ]
                        [                        
                            Prediction, text width=5em
                            [
                                PromptTrack~\cite{wu2023language}{, }BEV-InMLLM~\cite{ding2024holistic}, leaf, text width=51.3em
                            ]
                        ]
                    ]
                    [
                        Embodied AI \\
                        \ref{embodied}
                        [
                            Perception, text width=5.2em[
                                OpenEQA \cite{majumdar2024openeqa}{, }
                                AffordanceLLM \cite{qian2024affordancellm}{, }3DVisProg\cite{yuan2024visual}{, }3D-LLM \cite{hong20233d}{, }3DVG\cite{wang2024solving}{, }REPLAN\cite{skreta2024replan},leaf, text width=51.2em
                            ]
                        ]
                        [                        
                            Minipulation, text width=5.2em
                            [
                                RT-2~\cite{zitkovich2023rt}{, }
                                RT-X~\cite{open_x_embodiment_rt_x_2023}{, }PaLM-E\cite{driess2023palm}{, }Instruct2Act~\cite{huang2023instruct2act}{, }RoboFlamingo~\cite{li2024roboflamingo}{, }
                                VoxPoser~\cite{huang2023voxposer}{, }LLARVA~\cite{niu2024llarva}{, }\\ManipLLM~\cite{li2024manipllm}{, }OpenVLA~\cite{kim2024openvla}, 
                                leaf, text width=51.2em
                            ]
                        ]
                        [
                            Planning, text width=5.2em
                            [
                                NLMap~\cite{chen2023open}{, }ELLM~\cite{du2023guiding}{, }LLaRP~\cite{szot2023large}{, }LL3DA~\cite{chen2024ll3da}{}, leaf, text width=51em
                            ]
                        ]
                        [
                            Navigation, text width=5em
                            [
                                LM-Nav \cite{shah2022lmnav}{, }MultiPLY~\cite{hong2024multiply}{, }EMMA~\cite{yang2024embodied}{, }Conceptgraphs~\cite{gu2023conceptgraphs}{, }RILA\cite{yang2024rila},
                                leaf, text width=51.3em
                            ]
                        ]
                    ]
                    [
                        Automated Tool\\ management
                        \ref{tool}, text width=8em
                        [MM-REACT\cite{yang2023mm}{, }AVIS\cite{hu2024avis}{, }MLLM-Tool\cite{wang2024tool}{, }HuggingGPT\cite{shen2024hugginggpt}{, }CLOVA\cite{gao2024clova}{, }TaskMatrix.AI\cite{liang2024taskmatrix}{, }ViperGPT\cite{suris2023vipergpt}{, }\\ GPT4Tools\cite{yang2024gpt4tools}{, }CRAFT\cite{yuan2023craft}{, }ToolkenGPT\cite{hao2024toolkengpt}{, }Chameleon\cite{lu2024chameleon}{, }TroVE\cite{wang2024trove}{, }Mind's Eye~\cite{liu2022mind}{, }Confucius\cite{gao2024confucius}{, }Agent-S\cite{agashe2024agent}{, }\\AgentStore\cite{jia2024agentstore}{, }OS-Atlas\cite{wu2024atlas}{, }LogiCode\cite{zhang2024logicode}{, }ICAL\cite{zhang2024logicode}{, }ExACT\cite{yu2024exact}{, }Agent-Occam\cite{yang2024agentoccam}{, }Learn-by-interact~\cite{anonymous2024learnbyinteract}, leaf, text width=56em]
                    ]
                ]
                [
                    Text-to-vision\\
                    \ref{text2vision}
                    [
                        Text to image \\
                        \ref{text2image}
                        [GILL\cite{koh2024generating}{, }LLMGA \cite{xia2023llmga}{, }LLM-grounded Diffusion \cite{lian2024llmgrounded}{, }SEED \cite{ge2023planting}{, }LaVIT \cite{jin2023unified}{, }Emu \cite{sun2023generative}{, }VL-GPT \cite{zhu2023vlgpt}{, }MiniGPT-5 \cite{zheng2023minigpt}{,}\\Kosmos-G \cite{pan2023kosmos}{, }CoDi-2 \cite{tang2023codi}{, }JAM \cite{aiello2023jointly}{, }CM3Leon \cite{yu2023scaling}{, }DiffusionGPT \cite{qin2024diffusiongpt}{, }SEED-LLaMa \cite{ge2023making}\\, leaf, text width=58.5em]
                    ]
                    [
                        Text to 3D \\
                        \ref{text23D}
                        [LI3D \cite{lin2023towards}{, }3D-GPT \cite{sun20233d}{, }ShapeGPT \cite{yin2023shapegpt}{, }Holodeck \cite{yang2024holodeck}{, }LLMR \cite{de2023llmr}{, }MeshGPT \cite{siddiqui2023meshgpt}{, }GPT4Point \cite{GPT4Point}\\, leaf, text width=58.5em]
                    ]
                    [
                        Text to video \\
                        \ref{text2video}
                        [LVD \cite{lian2024llmgrounded}{, }VideoDrafter \cite{long2024videodrafter}{, }Free-bloom \cite{huang2023free}{, }VideoDirectorGPT \cite{lin2023videodirectorgpt}{, }Mora \cite{yuan2024mora}{, }SVD \cite{blattmann2023stable}{,} MAGVITv2 \cite{yu2023language}{, }VideoPoet \cite{kondratyuk2023videopoet}, leaf, text width=58.5em]
                    ]
                ]
            ]
        \end{forest}
        }
    \centering
    \caption{VLLM application texonomy. REC (Referring Expression Comprehension), RES (Referring Expression Segmentation), OCR (Optical Character Recognition).}
    \label{fig:mindmap of ML foundation}
\end{figure*}

%% file: contents/image2text.tex
\subsection{Image-to-text}\label{img2txt}
\begin{figure*}
    \centering
    \includegraphics[width=0.81\linewidth]{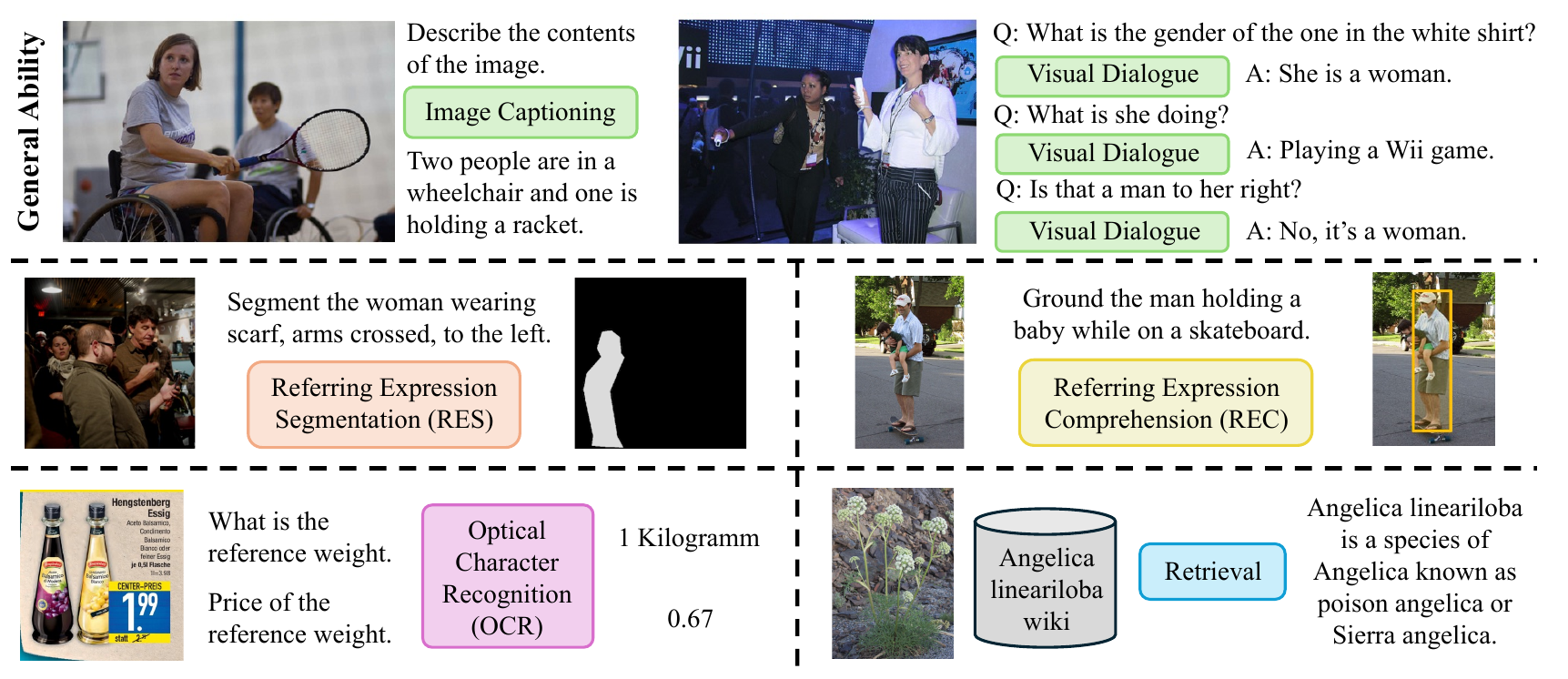}
    \caption{An illustration of image-to-text general domain applications.}
    \label{fig:img2text_general}
\end{figure*}
Image-to-text tasks are fundamentally designed for models to extract image features and translate them into human language. Traditionally, methods such as CNNs combined with RNNs are the dominant solution for this task. These models primarily focus on leveraging the spatial information of images to generate corresponding textual descriptions. However, such methods demonstrate limitations in handling the complexity and variability of real-world images. Recently, with the rise of LLMs, numerous studies have explored leveraging the power of these advanced models for image understanding tasks. In particular, when combined with a robust pretrained vision encoder, such as CLIP-based or CNN-based architectures, these models have significantly enhanced the ability to  accurately and contextually interpret and describe visual content. These advancements have enabled a wide range of applications, including general domain tasks such as image captioning, visual question answering (VQA), visual dialogue, referring expression comprehension (REC), referring expression segmentation (RES), and optical character recognition (OCR), as well as applications in specific domains.

\subsubsection{General domain application} In the category of general domain applications, image-to-text models are designed to address multiple visual understanding tasks, rather than focusing on any specific domain. Traditionally, most computer vision models were developed to solve single, discrete tasks such as classification, detection, or segmentation. However, these models were limited in providing generalist solutions capable of handling multiple vision tasks simultaneously. Recently, in the field of natural language processing, the instruction-tuning paradigm introduced in some large language models (LLMs) \cite{raffel2020exploring, brown2020language} has proven to be a game-changer, ushering in a new era where LLMs can manage a diverse range of text-based tasks. This paradigm shift has also influenced the vision domain, leading to the development of instruction-tuned visual LLMs capable of effectively interpreting and describing visual content across various tasks, such as image captioning and VQA. Numerous studies \cite{liu2024visual, liu2023improved, dai2024instructblip, zhu2023minigpt, chen2023minigpt, ye2023mplug} have pioneered the use of visual instruction tuning, achieving superior performance on these tasks and demonstrating its potential to enable more versatile and generalizable visual understanding models.

\textbf{General Ability.}
Image captioning, VQA and visual dialogue are fundamental capabilities of most vision-language image-to-text models. Captioning involves generating descriptive textual explanations for visual data, while VQA focuses on answering specific questions. Visual dialogue further extends VQA by enabling multi-turn conversations about the visual content. Currently, nearly all visual LLMs employ multiple training stages to develop these capabilities.

The first stage, known as pretraining, involves training models on extensive datasets of image-text pairs, such as COCO \cite{lin2014microsoft}, Visual Genome \cite{krishna2017visual}, CC3M \cite{changpinyo2021conceptual}, Conceptual Captions \cite{sharma2018conceptual}, and LAION400M \cite{schuhmann2021laion}, \textit{etc}. The primary objective during this phase is feature alignment, which unifies image and text into a single embedding space. This alignment process ensures that the model can understand and process visual and textual information cohesively.

During pretraining, connectors between the vision encoder and LLMs are trained to synchronize visual and language representations.  These connectors include linear projectors \cite{liu2024visual, chen2023minigpt}, MLPs \cite{liu2023improved, lu2024deepseek}, Q-formers \cite{dai2023instructblip, zhu2023minigpt}, and Resamplers \cite{ye2023mplug, bai2023qwen}, \textit{etc}. These connectors serve as bridges aligning the visual and language spaces, working in conjunction with the pretraining objective of the image captioning task. This alignment enhances the models' efficiency and effectiveness, enabling them to excel in generating coherent textual descriptions from visual inputs.

In the subsequent training stage, referred as supervised fine-tuning, models are trained on high-quality image-text pair datasets curated from high-capacity closed-source resources like GPT-4V \cite{achiam2023gpt}, or Gemini~\cite{team2023gemini}. Examples include LLaVA-Instruct-158K \cite{liu2024visual}, GRIT \cite{peng2023kosmos}, ShareGPT4V \cite{chen2023sharegpt4v}, and LVIS-Instruct4V \cite{wang2023see}. Additionally, some studies \cite{ye2023mplug2} integrate high-quality text-only dialogue datasets, including ShareGPT-80K \cite{chiang2023vicuna}, SlimOrca \cite{SlimOrca}, the Alpaca dataset \cite{alpaca}, and the Vicuna dataset \cite{vicuna2023}. These datasets help to preserve the model's ability to handle multi-turn human dialogues. This fine-tuning stage equips models with the ability to  effectively follow user instructions and enhances their multimodal conversation capabilities, thereby improving their performance in visual dialogue tasks.

During fine-tuning, some models \cite{liu2024visual, bai2023qwen, zhang2023internlm} opt to freeze the vision encoder to preserve the prior knowledge of visual representations. In contrast, others \cite{ye2023mplug2, li2024monkey} train both the vision encoder and LLMs to enhance the model's robustness in handling complex visual input. Furthermore, additional training stages are integrated into the training pipeline of certain models. For example, MiniGPT-v2 \cite{chen2023minigpt}, Qwen-VL \cite{bai2023qwen}, and DeepSeek-VL \cite{lu2024deepseek} incorporate extra training stages to strengthen their conversational capabilities.

\textbf{Referring Expression Segmentation (RES).} Object segmentation is a fundamental task in computer vision, involving the localization of objects at the pixel level. Traditional methods have addressed this task by targeting a predefined set of objects. Recently, the open vocabulary approach has gained prominence, leveraging LLMs to segment objects based on natural language descriptions, known as \textit{referring expressions}.

LISA \cite{lai2024lisa} introduces the novel task of \textit{Reasoning Segmentation}, where models generate segmentation masks guided by complex and implicit query text instructions. This architecture integrates a VLLM with an additional vision backbone, specifically the Segment Anything Model (SAM) \cite{kirillov2023segment}, to extract visual embeddings. Leveraging an embedding-as-mask paradigm, the model decodes the last-layer embedding of the \texttt{<SEG>} token from the VLLM into a segmentation mask, thereby enhancing its ability to address intricate reasoning and leverage world knowledge effectively.

GSVA \cite{xia2024gsva} builds on LISA by introducing multiple \texttt{<SEG>} tokens to handle multiple targets and \texttt{<REJ>} tokens to reject empty targets, effectively addressing the challenges of Generalized Referring Expression Segmentation (GRES). Several other works, such as Osprey \cite{yuan2024osprey}, VisionLLMv2 \cite{wu2024visionllm}, GLaMM \cite{rasheed2024glamm}, and GROUNDHOG \cite{zhang2024groundhog}, integrate mask or region extractors to enhance the model's ability to capture fine-grained details. This strategy significantly improves RES accuracy, highlighting the effectiveness of incorporating specialized extractors for managing complex visual details.

Unlike LISA and other approaches that use SAM \cite{kirillov2023segment} as a supporting segmentation model, VistaLLM \cite{pramanick2024jack} and LLaFS \cite{zhu2024llafs} directly generate segmentation coordinates from the LLM decoders. These models simplify the segmentation process by leveraging the LLMs to directly generate precise segmentation coordinates, eliminating the need for an intermediate segmentation model.

\textbf{Referring Expression Comprehension (REC).} REC is a critical computer vision task that involves identifying and localizing objects in an image based on natural language descriptions. Early works like Shikra \cite{chen2023shikra} and Kosmos-2 \cite{peng2023kosmos} build on the architectures of LLaVA \cite{liu2024visual} and Kosmos-1 \cite{huang2023language}, using instruction-tuning datasets for grounding and referring tasks. These models represent location coordinates as text tokens, which are generated by LLMs.

Similarly, VisionLLM \cite{wang2024visionllm}, Pink \cite{xuan2024pink}, ChatSpot \cite{zhao2023chatspot}, InfMLLM \cite{zhou2023infmllm}, and ASMv2 \cite{wang2024all} aim to enhance training datasets by incorporating more region-level, fine-grained information. Models such as Ferret \cite{you2023ferret}, GPT4RoI \cite{zhang2023gpt4roi}, Ferret-v2 \cite{zhang2024ferret}, and Lion \cite{chen2024lion} modify their architectures to better capture detailed features from input images. Specifically, Ferret and GPT4RoI leverage Spatial-Aware Visual Sampler and Region Feature Extractor modules, allowing these models to focus on and extract critical image regions, thereby improving grounding accuracy.

Sphinx \cite{lin2023sphinx} leverages multiple image encoders, including CLIP-ViT \cite{radford2021learning}, CLIP-ConvNeXt \cite{liu2022convnet}, and DINOv2 \cite{oquab2023dinov2}, and introduces a novel joint mixing strategy to integrate model weights. This approach enhances multi-modal understanding and delivers superior performance in fine-grained visual tasks.

An alternative approach leverages external modules for precise object localization instead of directly generating bounding boxes from LLMs. For instance, DetGPT \cite{pi2023detgpt} extracts all objects using the GroundingDINO \cite{liu2023grounding} model for the grounding task. BuboGPT \cite{zhao2023bubogpt} introduces a pipeline that combines the Recognize Anything Model (RAM) \cite{zhang2024recognize}, GroundingDINO, and the SAM \cite{kirillov2023segment} to align bounding boxes with objects in the output sequence.

Further, Lenna \cite{wei2023lenna} and VisionLLMv2 utilize hidden features of the \texttt{<DET>} token as input for the detection module, following GroundingDINO’s architecture. LLaVA-Grounding \cite{zhang2023llava} employs OpenSeeD \cite{zhang2023simple} as its grounding model, using the \texttt{<REG>} token to enhance detection processes.

Recently, OMG-LLaVA \cite{zhang2024omg} introduces a new architecture that includes a universal perception module combined with an LLM. This enables unified image-level, object-level, and pixel-level reasoning and understanding tasks, achieving state-of-the-art performance on datasets such as refCOCO, refCOCO+, refCOCOg \cite{mao2016generation}, and GranD \cite{rasheed2024glamm}.

\textbf{Optical Character Recognition (OCR).} OCR is a crucial application within image-to-text tasks, aiming to convert various types of documents—such as scanned paper documents, PDFs, or images captured by digital cameras—into editable and searchable data. With the advent of LLMs, particularly VLLMs, users can now interact more effectively with text-rich documents. Models like LLaVAR \cite{zhang2023llavar}, mPLUG-DocOwl \cite{ye2023mplug}, UniDoc \cite{fengunidoc}, and TGDoc \cite{wang2023towards} are pioneering in this direction. These models build on the robust architectures and training procedures of LLaVA \cite{liu2024visual}, mPLUG-Owl \cite{ye2023mplug}, MiniGPT-4 \cite{zhu2023minigpt}, and Shikra \cite{chen2023shikra}, respectively, and have been further trained using instruction-tuning datasets comprising a diverse range of text-rich images, including posters, book covers, and infographics.

One of the key techniques that enhance a model's capability in OCR-related tasks is the ability to handle high-resolution data. Many of the previous models typically leverage pretrained CLIP \cite{radford2021learning} as the vision encoder, which has limited input resolutions of 224 or 336 pixels. To overcome this limitation, UReader \cite{ye2023ureader} employs a shape-adaptive cropping module that segments high-resolution images into smaller regions, enabling effective use of low-resolution vision encoders. Conversely, DocPedia \cite{feng2023docpedia} takes a novel approach by direcly handling high-resolution inputs in the frequency domain, capturing richer visual and textual information while requiring fewer visual tokens.

TextMonkey \cite{liu2024textmonkey} leverages the Shifted Window Attention mechanism in the Swin Transformer \cite{liu2021swin} to preserve cross-window connectivity and employs a token resampler to filter relevant tokens, enhancing performance across various text-centric tasks. DocKylin \cite{zhang2024dockylin} introduces Adaptive Pixel Slimming (APS), which leverages gradient information to eliminate redundant visual content, and Dynamic Token Slimming (DTS) modules to filter out irrelevant visual tokens, thereby improving the efficiency and effectiveness of visual document understanding. LayoutLLM \cite{luo2024layoutllm} integrates layout instruction tuning by combining layout-aware pre-training with supervised fine-tuning, enabling it to effectively capture and utilize document layout information.

Similarly, Fox \cite{liu2024focus} focuses on fine-grained, multi-page document understanding by employing flexible, position-aware prompts and hybrid data synthesis, efficiently leveraging multiple vision vocabularies. HRVDA \cite{liu2024hrvda} employs content filtering and instruction filtering modules to eliminate content-agnostic and instruction-agnostic visual tokens, optimizing model training and inference for high-resolution images. More recently, InternLM-XComposer2-4KHD \cite{dong2024internlm} has pushed the boundaries by handling resolutions up to 4K HD. It employs dynamic resolution and automatic patch configuration to accommodate varying resolutions and aspect ratios, achieving state-of-the-art performance across OCR benchmarks such as DocVQA, ChartQA, InfoVQA, and TextVQA.

\textbf{Retrieval.} Retrieval tasks in VLLMs focus on selecting and retrieving relevant images or textual data based on a given query, which can be either visual or textual in nature. These tasks are fundamental to the effectiveness of image-to-text models, enabling accurate and context-aware outputs by bridging the gap between visual and textual modalities. Recent advancements in retrieval-based approaches can be categorized into two main groups: single-stage retrieval models and hierarchical retrieval-augmented models.
\begin{itemize}
     \item \textit{Single-Stage Retrieval Models} are characterized by their capability to perform direct retrieval operations by mapping visual and textual data into a unified embedding space. These models typically employ a straightforward retrieval process, where the model selects the most relevant image or text based on learned embeddings. For instance, FROMAGe \cite{koh2023grounding} pioneered this approach by integrating vision and language through the use of a specialized retrieval token ([RET]), which enhances the model's ability to accurately retrieve images and generate text. UNIMUR \cite{wang2024unified} further refines this concept by introducing a unified multimodal embedding training paradigm, addressing modality bias and ensuring more consistent and coherent retrieval results across both visual and textual domains.

    \item \textit{Muli-Stage Retrieval Models} are designed to enhance retrieval tasks by incorporating multiple layers of retrieval processes, often leveraging external knowledge sources to enrich the contextual relevance of the outputs. These models typically involve a two-step retrieval process, where the most relevant documents or passages are first identified and then used to generate more accurate and contextually rich responses. For example, Wiki-LLaVA \cite{caffagni2024wiki} employs a hierarchical retrieval process. It first locates relevant documents using visual and textual encoders, then extracts pertinent passages within those documents. This approach enriches the input context, particularly for tasks requiring external knowledge, such as visual question answering. EchoSight \cite{yan2024echosight} extends this approach by integrating multimodal inputs, combining visual perception with acoustic signals, to retrieve and synthesize information across multiple modalities, thereby enhancing the model’s ability to handle complex, multimodal queries.

\end{itemize}

\subsubsection{Specific Domain Application}
\begin{figure*} 
  \centering
  \includegraphics[width=\textwidth]{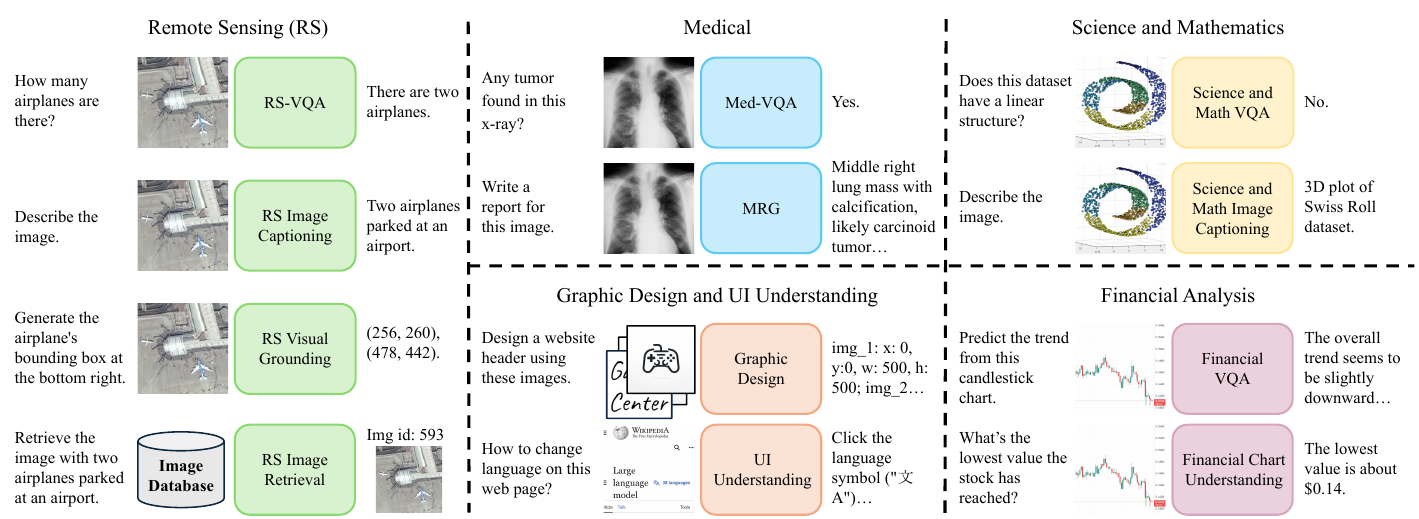}
  \caption{An illustration of image-to-text specific domain applications.}
  \label{fig:img2txt}
\end{figure*}

The advancement of VLLM also paves a new way for image-to-text applications across various specialized domains. Unlike VLLMs designed for general applications as discussed in the previous section, these domain-specific models need extensive training on specialized datasets or receive unique structural modifications to improve performance. By aligning the models with the intricate demands of each domain, they demonstrate significant performance improvements, outperforming general models in specific tasks. This high level of contextual relevance and model performance are essential for addressing the unique challenge of many domains, which include:

\textbf{Remote Sensing (RS)}. The inherent differences between natural and RS images, including variations in color distributions, image resolutions, image context, and object scales, present significant challenges for using general VLLMs in tasks within the RS domain. These tasks include RS captioning \cite{rs16091477}, \cite{silva2024large}, \cite{zhang2024earthgpt},\cite{zhan2024skyeyegpt}, \cite{kuckreja2023geochat}, \cite{muhtar2024lhrsbot}, RS visual grounding \cite{pang2024h2rsvlm}, \cite{zhang2024popeye},\cite{muhtar2024lhrsbot}, \cite{zhang2024earthgpt}, \cite{zhan2024skyeyegpt}, \cite{kuckreja2023geochat}, RS-VQA \cite{rs16091477}, \cite{pang2024h2rsvlm}, \cite{muhtar2024lhrsbot},\cite{zhang2024earthgpt}, \cite{zhan2024skyeyegpt}, \cite{kuckreja2023geochat}, \cite{hu2023rsgpt}, and RS image retrieval \cite{silva2024large}. To address these challenges, numerous studies concentrate on fine-tuning VLLMs for RS tasks by developing specialized RS datasets.

RS-LLaVA \cite{rs16091477} constructs a tailored RS-instruction dataset by integrating data from public datasets on the RS captioning and RS-VQA domains. GeoChat \cite{kuckreja2023geochat} incorporates its dataset by incorporating public datasets on different tasks, including RS visual grounding and landcover classification, and proposes a dataset specifically for flood detection. RSGPT \cite{hu2023rsgpt} manually annotates a high-quality RS captioning dataset enriched with detailed scene descriptions and comprehensive object information. SkyEyeGPT \cite{zhan2024skyeyegpt} aims at unifying diverse RS image-to-text tasks, encompassing both single-task and multi-task conversational instructions, with manual verification to ensure dataset quality. EarthGPT \cite{zhang2024earthgpt} overcomes the limitation of existing models, which mainly train on optical RS images by incorporating multiple types of RS images, such as synthetic aperture radar (SAR) and infrared images, into their dataset. This integration enhances the model's performance across various RS tasks. Similarly, LHRS-Bot \cite{muhtar2024lhrsbot} further enriches the instruction dataset by utilizing extensive volunteered geographic information (VGI) and globally available RS images from Google Earth, improving model performance across most of the RS tasks. H\textsuperscript{2}RSVLM \cite{pang2024h2rsvlm} includes a range of unanswerable questions into their RS-VQA dataset, which is then used to fine-tunes LLaVA \cite{liu2024visual} to successfully mitigate hallucination issue.

Beyond general RS tasks, some researchers create specialized datasets to tackle specific RS challenges. For example, Silva \textit{et al}. \cite{silva2024large} utilize various RS image captioning datasets and train linear layers to apply contrastive learning between RS images and special tokens to address RS image retrieval tasks. Popeye \cite{zhang2024popeye} focuses on ship detection by constructing a dataset that merges data from four existing ship detection datasets effectively targeting this specific issue.

\textbf{Medical}. The integration of VLLMs into the medical domain represents a substantial advancement in diagnostics and treatment planning. These models leverage VQA to interpret complex medical images \cite{zhou2024generalist}, \cite{li2023llavamed}, \cite{moor2023medflamingo}, \cite{zhang2023pmcvqa}, \cite{zhou2025training}, and generate comprehensive and precise medical reports \cite{he2024pefomed}, \cite{thawkar2023xraygpt}, \cite{ranjit2023retrieval}, \cite{pellegrini2023radialog}.

In the medical field, Medical Visual Question Answering (Med-VQA) is a fundamental task to enhance diagnostic capabilities by analyzing various types of medical images such as CT scans, radiographs, and dermoscopy images using VLLMs. Med-VQA involves practitioners interacting with VLLMs to extract specific details from medical images, facilitating interactive diagnostic support. For example, MedVersa \cite{zhou2024generalist} demonstrates proficiency in handling multifaceted medical image interpretation by leveraging multimodal inputs and dynamic task specifications. PMC-VQA \cite{zhang2023pmcvqa} introduces a large-scale Med-VQA dataset encompassing diverse image modalities such as X-ray, MRI, and Computed Tomography, along with a broad spectrum of diseases. This model employs visual instruction tuning on this comprehensive dataset to enhance its capability to address a wide range of medical questions effectively. LLaVA-Med \cite{li2023llavamed} leverages a two-step fine-tuning process based on LLaVA \cite{liu2024visual}: first, biomedical vocabulary training using image-caption pairs, followed by medical instruction training utilizing GPT-4 \cite{achiam2023gpt} generated instruction-following data. Remarkably, this approach requires only one day of training with eight A100 GPUs to achieve state-of-the-art performance in Med-VQA benchmarks \cite{liu2021slake}, \cite{he2020pathvqa}, \cite{lau2018}. Med-Flamingo \cite{moor2023medflamingo} also demonstrates the effectiveness of few-shot learning in generating accurate responses to complex medical queries. Furthermore, Zhou et al.~\cite{zhou2025training} propose a two-stage fine-tuning strategy incorporating instruction tuning and reinforcement learning that allows the model to focus more effectively on abnormal regions in medical images, outperforming other open-source models on the abnormality recognition task.

Building on the advancements in Med-VQA, the task of Medical Report Generation (MRG) further streamlines the diagnostic workflow by automating the creation of comprehensive medical reports. While Med-VQA leverages VLLMs to extract specific details and provide immediate diagnostic support through interactions, MRG augments this by generating detailed, structured reports highlighting anomalies and suggesting potential diagnoses using VLLMs. For example, PeFoMed \cite{he2024pefomed} introduces a novel parameter-efficient fine-tuning framework for MRG tasks, showing significant improvements in generating clinically useful reports. XrayGPT \cite{thawkar2023xraygpt} enhances the quality and usability of automated radiology reports by training on high-quality visual and textual synthetic summaries from free-text radiology reports as training data. Ranjit \textit{et al.} \cite{ranjit2023retrieval} employ the Retrieval Augmented Generation (RAG), integrating domain-specific retrievals with general language models to generate accurate and clinically relevant radiology reports, while reducing hallucinations and irrelevant content. RaDialog \cite{pellegrini2023radialog} achieves state-of-the-art results in generating and interactively modifying radiology reports, facilitating expert collaboration and enhanced diagnostic support.

\textbf{Science and Mathematics. } Beyond applications in Medical and RS domains, researchers also integrate VLLMs into the science domain to enhance understanding of diagrams and graphs in academic settings. Science VQA \cite{lu2022learn}, \cite{wang2023tsciq}, \cite{hu2024mplugpaperowl} is a prevalent evaluation task in these integrations. Specifically, Lu \textit{et al.} \cite{lu2022learn} introduce a large-scale science VQA dataset, ScienceQA, encompassing diverse topics such as natural science, social science, and language science. T-SciQ \cite{wang2023tsciq} develops a pipeline for generating high-quality Chain-of-Thought (CoT) prompts to fine-tune VLLMs, achieving state-of-the-art performance on the ScienceQA benchmark. mPLUG-PaperOwl \cite{hu2024mplugpaperowl} fine-tunes mPLUG-DocOwl \cite{ye2023mplugdocowl} on a scientific diagram analysis dataset derived from 48,000 high-quality arXiv papers on Machine Learning (ML), addressing both Science VQA and science image captioning tasks. FigurA11y \cite{figura11y} further explores the science image captioning task by designing a system, based on GPT-4 \cite{achiam2023gpt} and LLaVA \cite{liu2024visual} to generate high-quality alternative text for scientific figures. This system aims to provide more detailed image descriptions, making scientific content more accessible to blind and low-vision readers.

Within the broader field of science, visual mathematics has emerged as a distinct domain, with many advancements significantly enhancing the ability of VLLMs to address complex mathematical problems and introduce new evaluation metrics. G-LLaVA \cite{gao2023gllava} introduces the Geo170K dataset \cite{gao2023gllava}, containing 170,000 geometry image-caption and question-answer pairs, to fine-tune LLaVA \cite{liu2024visual}, outperforming GPT-4V \cite{achiam2023gpt} in geometric mathematics VQA task. Similarly, Math-LLaVA \cite{shi2024mathllava} fine-tunes LLaVA 1.5 \cite{liu2024visual} with the MathV360K dataset \cite{shi2024mathllava}, which includes 40,000 diagrams and 360,000 question-answer pairs, achieving performance on par with GPT-4V in mathematical reasoning tasks. MAVIS \cite{zhang2024mavis} selects MAmmoTH2 model \cite{yue2024mammoth2} as the base model and fine-tunes it over the MAVIS-Instruct 834K dataset \cite{zhang2024mavis}, containing diagram, question-answer pairs, and rationale, achieving state-of-the-art results. MAmmoTH-VL \cite{guo2024mammothvl} Beyond model improvements, significant progress has been made in creating new benchmarks to evaluate the performance of VLLMs in solving visual mathematics problems. The MathVista benchmark \cite{lu2024mathvista} consolidates over 6,000 examples from 31 datasets, providing a comprehensive evaluation of the mathematical VQA task. GeoEval \cite{zhang2024geoeval} focuses on geometry mathematics questions, encompassing basic shapes, three-dimensional objects, and analytic geometry. CMMaTH \cite{li2024cmmath} establishes the largest Chinese multimodal mathematical benchmark, specifically for K12 math questions. Additionally, MathVerse \cite{zhang2024mathverse} and WE-MATH \cite{qiao2024wemath} introduce benchmarks that include crucial reasoning steps to evaluate the quality of CoT reasoning by VLLMs.

\textbf{Graphic Design and UI Understanding. } In our progressively digital world, graphical designs play a crucial role across a range of digital devices. Several studies successfully utilize VLLMs in graphics-related tasks, encompassing graphic design and UI understanding. For graphic design tasks, Graphist \cite{cheng2024graphic} is the first layout generation VLLM, that constructs graphic compositions from unordered sets of design elements formatted as a JSON draft protocol. This protocol details the coordinates, size, and sequence of each component. For user interface (UI) understanding, general VLLMs may fall short because UI images typically contain smaller objects of interest, such as icons and text, which differ significantly from those in natural images. To address this challenge, CogAgent \cite{hong2023cogagent} employs low and high-resolution image encoders to accurately recognize small page elements and text. Similarly, Ferret-UI \cite{you2024ferretui} divides the image into two sub-images and encodes them separately before feeding them into VLLMs, enhancing the model's ability to process intricate UI details.

\textbf{Financial Analysis. }Researchers also explore how VLLMs can be used in financial analysis to interpret financial charts and deliver valuable insights. FinVis-GPT~\cite{wang2023finvisgpt} develops a financial dataset featuring candlestick and line charts of Chinese A-shares, which is then used to fine-tune VLLMs for financial analysis. FinTral~\cite{bhatia2024fintral} adopts a similar approach, using stock price data from Fortune 500 companies to facilitate financial analysis in English contexts. FinTral achieves state-of-the-art performance in the Stock Movement Prediction task, outperforming other general VLLMs such as GPT-4V \cite{achiam2023gpt}, Gemini-Pro~\cite{geminiteam2024gemini}, and LLaVA-NEXT~\cite{liu2024llavanext}.

%% file: contents/video2text.tex
\subsection{Video-to-text}\label{video2text}
\begin{figure*}
    \centering
    \includegraphics[width=0.98\linewidth]{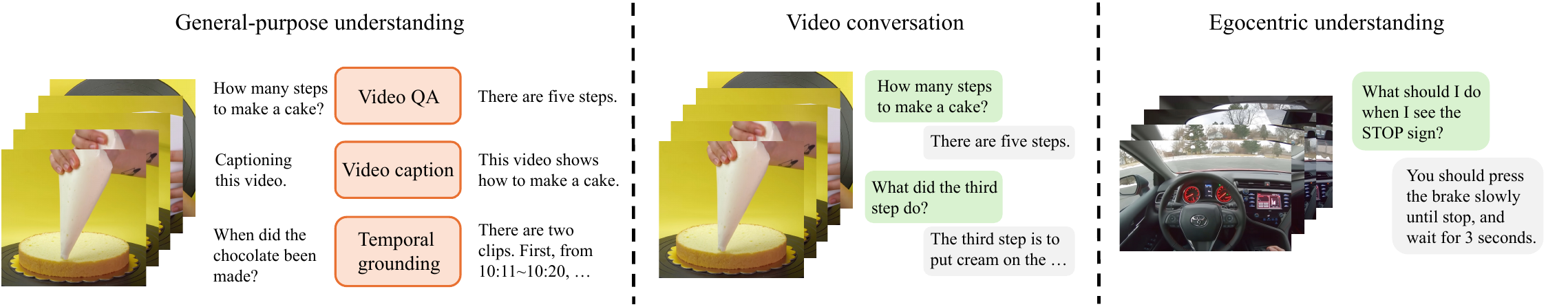}
    \caption{An illustration of video-to-text applications. }
    \label{fig:vid2text}
\end{figure*}

Video-to-text tasks deal with advanced video content understanding problems such as video captioning, video question answering, video conversation, \emph{etc}. The input videos could be either short within a few seconds, or a couple of hours long. They are accessed from historically cached databases or real-time streaming platforms. Compared to image-to-text tasks, video-to-text tasks are significantly more challenging because of the big semantic gap between sparse video content and abstract language information. In the spirit of LLMs, how to embed sparse causal information from long videos, \emph{e.g.}, temporal activity evolution, and physical motion dynamics, into a dense and tokenizable space for learning a video LLM is an open-ended research problem. According to the application goals, we categorize existing literature in the following parts.

\subsubsection{General-purpose Video Understanding} In the era of LLMs, textual language is becoming a unified interface for modeling various video understanding problems. Such a trend leads to many existing general-purpose video LLMs. The basic idea of building a general-purpose model is to perform large-scale model pre-training on billions or trillions of visual-text data. The pre-training stage brings generalizable knowledge for downstream video understanding tasks. For example, early work such as VideoBERT~\cite{videobert} stands on top of the pre-trained BERT~\cite{bert} to pre-train on large-scale video datasets by masked language modeling (MLM), which is found critical to both video generation and classification tasks. Instead of MLM, recent LLMs~\cite{lavila,vid2seq,zhang2023video,li2023llama,ataallah2024minigpt4,zhang2024llavanextvideo} that adopt autoregressive modeling (ARM) show much better capability, which aims to predict the next token in language space. 

The ``next-token prediction'' has inspired recent general-purpose video LLMs. For example, LaViLa~\cite{lavila} treats video data as a condition and adopts GPT2~\cite{gpt2} and T5-large~\cite{t5} as video narrator and rephrase, to fine-tune LLMs for applications including activity recognition, event retrieval, and video question answering. Vid2Seq~\cite{vid2seq} leverages large-scale narrated videos and time tokens in pre-training, leading to a generalizable model for multiple video captioning benchmarks. VideoLLaMA~\cite{zhang2023video} leverages video Q-former and ImageBind to handle visual temporal change and the modality correspondence between modalities of audio, video, and text. 
Following similar model architecture, recent video LLMs such as LLaMA-VID~\cite{li2023llama}, MiniGPT4-Video~\cite{ataallah2024minigpt4}, ST-LLM~\cite{liu2024st}, LLaVA-Next-Video~\cite{zhang2024llavanextvideo}, LLaVA-OV~\cite{li2024llava-ov}, and Aria~\cite{aria} have been introduced rapidly. 


\textbf{Long-form video LLMs.} The common idea of existing video LLM is to use a pre-trained LLM as the decoder to generate text responses. This raises a major challenge of how to effectively encode long-form videos into visual tokens and project them into text token space for LLM to decode. Recently, LLaMA-VID~\cite{li2023llama} uses two tokens including the context token and content token to encode each frame, enabling hourly-long video understanding. LVChat~\cite{wang2024lvchat} dynamically adjusts the number of video tokens by considering the duration of input videos. 
MA-LMM~\cite{he2024ma} stores the features of historical video tokens in a memory bank such that the model can process hourly-long videos in an online manner. Different from previous methods, recent LongVU~\cite{shen2024longvu} adopts temporal frame selection and spatial token selection to significantly reduce redundant tokens in video LLMs. Until today, how to efficiently encode long video-text context into a unified token space for video conversation is still challenging and being actively studied in research communities. 

\subsubsection{Video Conversation} Instead of building foundational video-to-text models for general purposes, video conversation raises a great interest because the language model serves as a conversational agent of humans. However, this scenario naturally requires multiple rounds of reasoning steps to understand the complex video content, as well as humans' ambiguous intent in a long dialog context. This leads to the recent trend of studying video-language chat applications. ChatGPT is generally regarded as a breakthrough that successfully makes LLMs such as GPT-4~\cite{openai2023gpt} applicable for complex reasoning tasks. Recently, its multi-modality version GPT-4(V)~\cite{gpt4v} takes images and video into account in conversation. These agents are commercial products without public code and models released. 
Since LLaVA~\cite{liu2024visual} is a representative visual assistant for image-to-text agent applications. Thanks to its simplicity, plenty of video assistants are developed recently, such as the Video-LLaVA~\cite{lin2023video}, VideoChat~\cite{li2023videochat}, VideoChat2~\cite{li2023mvbench}, VideoChatGPT~\cite{maaz2023video}, Valley~\cite{luo2023valley}, BT-Adapter-LLaVA~\cite{liu2023one}. Note that most existing general-purpose video LLMs are technically feasible for video conversation, though they are not optimized from engineering aspects for improving the conversation experience.

\subsubsection{Egocentric Understanding} For video understanding tasks, most applications focus on the third-person view. 
However, understanding egocentric human activities is essential for plenty of real-world applications. For example, a headset with an egocentric camera sensing system could assist a worker in finishing complex mechanical repair jobs, when the system can understand the humans' egocentric visual world and interact with humans with plain language.  
Prior research such as \cite{lin2022egocentric,pramanick2023egovlpv2,pei2024egovideo,chatterjee2024opening,Shen_2024_CVPR,xu2024egonce++} primarily resort to vision-language models such as CLIP~\cite{xue2022clip} to achieve egocentric video understanding, which are too small (less than 1 billion parameters) to handle complicated video tasks. Powered by advances in LLMs, the recent GPT4Ego~\cite{dai2024gpt4ego} leverages ChatGPT and chain-of-thought to generate text prompts for egocentric videos, achieving remarkable zero-shot egocentric action recognition performance. By leveraging LLMs to generate video narrations, LAVILA~\cite{Zhao_2023_CVPR} achieves large performance gains on both egocentric and third-person video understanding tasks. Following prompt tuning strategy, PALM~\cite{kim2025palm} treats the past video events and visual tokens to fine-tune LLM for egocentric long-term action anticipation. Different from PALM, recent AntGPT~\cite{zhao2023antgpt} takes advantage of LLMs at the bottom-up next-token prediction and top-down task goal planning for egocentric long-term action anticipation. To unify the egocentric visual memory and robotic control, recent LLM-Brain~\cite{mai2023llm} treats LLM such as BLIPv2 as the central brain of the model to handle users' instructions for multiple embodied AI tasks. To bridge the gap between LLMs and egocentric video content, Wang et al~\cite{wang2023egocentric} formulates a natural language query (NLQ) benchmark for egocentric video understanding, and introduce a video captioning module before applying LLMs. Recently, action programming has been an important application of egocentric video understanding, which requires complex visual reasoning capability of models over an egocentric view. The LEAP~\cite{dessalene2023leap} employs LLMs to generate executable action programs from egocentric videos, which is critical for robot manipulation, planning, and navigation.



%% file: contents/vision2action.tex
\subsection{Vision-to-action}\label{vision2action}

\begin{figure*}
    \centering
    \includegraphics[width=0.9\linewidth]{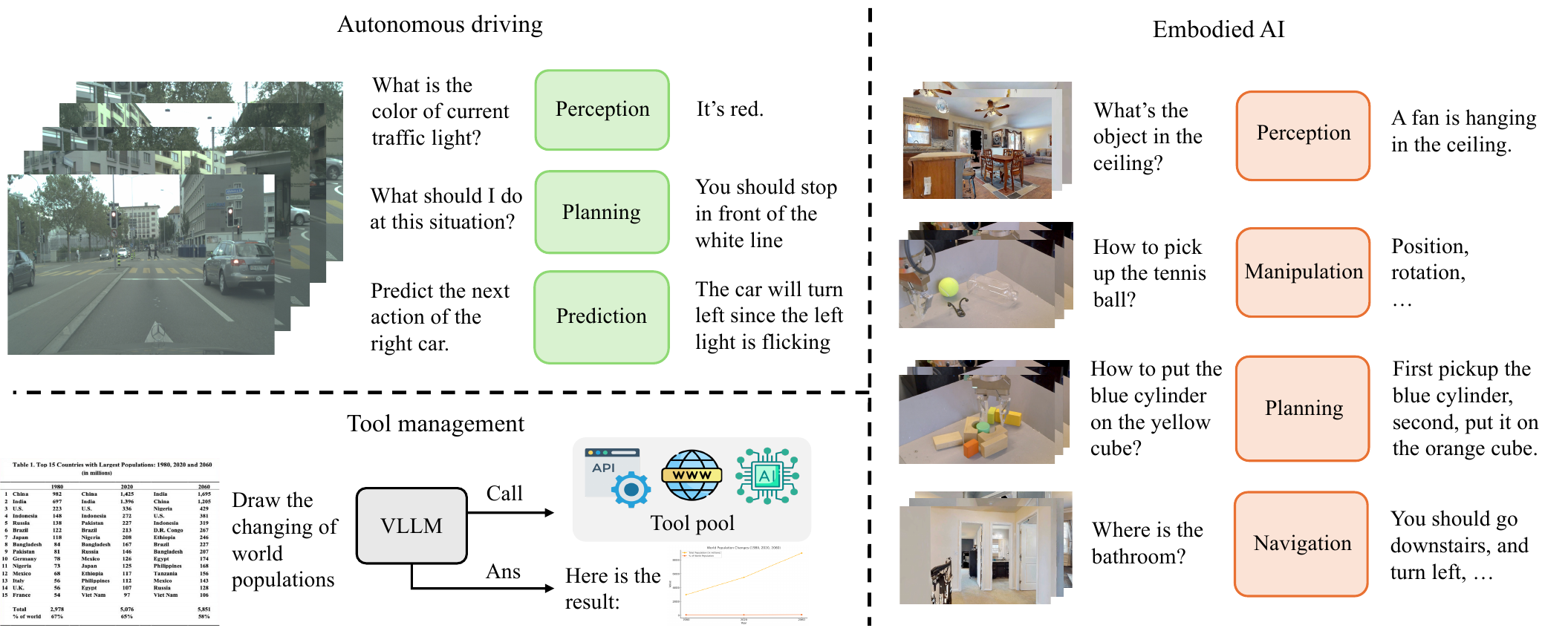}
    \caption{An illustration of vision-to-action applications.}
    \label{fig:vision2action}
\end{figure*}

Vision-to-action tasks based on VLLMs mainly take visual modalities such as images/videos/depth/3D as the visual input conditioned with language instructions, and the VLLM will generate actions to control behaviors of vehicles, robots or software (\textit{e.g.}, APIs), \textit{etc}. We divide such tasks into three main streams according to the application scenarios: autonomous driving (AD), embodied AI and tool management. These VLLM agents \cite{xie2024large} will be more intelligent and general when making decisions by equipping with LLM's strong contextualizing, reasoning, and generalization ability.

\subsubsection{Autonomous Driving}\label{autonomousdriving}
Recent AD systems transit from rule-based to data-driven ones \cite{yang2023llm4drive,cui2024survey,park2023visual}, and many of these methods resort to LLMs for better perception, planning and prediction ability. The vision input for LLMs in AD can be various and adaptive, like multi-view images \cite{tian2024drivevlm}, video \cite{xu2023drivegpt4}, bird's eye view (BEV) map \cite{dewangan2023talk2bev} and LiDAR \cite{yang2023lidar, shao2023lmdrive}, which is easy to be extended to various driving tasks. 

\textbf{Perception.} As mentioned in Sec. \ref{img2txt}, VLLMs have shown strong reasoning and zero-shot abilities in understanding tasks. This characteristic can also be extended to the AD system to improve the generalization capability in complex driving environments \cite{bojarski2016end}.
NuScences-QA \cite{qian2024nuscenes} proposes a VQA-based benchmark for the AD scenario, including both images and point clouds modalities as input and the question answering pairs on the existence, counting, status, types, and relationships of the objects. DriveLM \cite{sima2023drivelm} introduces a graph-structured reasoning task to model the logical dependencies during the AD question and answering process. Talk2BEV \cite{dewangan2023talk2bev} proposes to utilize the BEV map to perform scene understanding, visual reasoning, and free-form conversation tasks. HiLM-D \cite{ding2023hilm} proposes an efficient method to localize risk objects, predict intentions, and provide suggestions. Specifically, HiLM-D designs two branches for high-resolution perception and low-resolution reasoning, respectively. Built upon OpenFlamingo\cite{awadalla2023openflamingo}, Dolphins \cite{ma2023dolphins} can process data integrating video/images, instructions, and historical control signals, and generate the understanding results of the driving scenario. Specifically, a new CoT process (Grounded Chain-of-Thought, GCoT) is proposed to enhance the reasoning ability of Dolphins by providing the spatial information of the objects. LiDAR-LLM \cite{yang2023lidar} takes 3D LiDAR data as input and performs outdoor scene understanding like description, grounding and VQA based on a frozen LLM.

\textbf{Planning.} VLLMs can perceive complex driving scenarios and generate driving maneuvers or control signals for drivers based on current visual features.
ADriver-I \cite{jia2023adriver} introduces the interleaved vision-action pair as the inputs for autoregressively predicting the control signals of the current frame. It then employs a diffusion model to forecast future frames based on historical vision-action pairs, facilitating future action prediction. DriveVLM \cite{tian2024drivevlm} capitalizes VLLMs for scene understanding and planning  by containing a Chain-of-Thought (CoT) process, including scene description, scene analysis and hierarchical planning. DriveVLM can generate three kinds of driving plans, which are meta-actions for a short-term decision, decision description for articulating the more fine-grained driving strategy, and trajectory waypoints for depicting the vehicle's path. DiLu \cite{wen2023dilu} improves the planning capability of AD by combining a reasoning and reflection module based on past driving knowledge. Specifically, a memory module is employed to store the past driving experience like decision prompts, reasoning processes, \textit{etc.}, which can be utilized to enhance the reasoning and planning capabilities of the LLMs.
DriveGPT4 \cite{xu2023drivegpt4} can predict the low-level control signals by tuning an end-to-end VLLM based on the BDD-X dataset \cite{kim2018textual} annotated by ChatGPT. Specifically, DriveGPT4 can analyze video frames by a pretrained vision encoder, and is also trained for two stages like LLaVA \cite{liu2024visual}. SurrealDriver \cite{jin2023surrealdriver} proposes a driver agent simulation framework for urban scenarios, which can provide driving maneuvers based on real world driver experience. LMDrive \cite{shao2023lmdrive}
 introduces an end-to-end driving framework, which can perceive multi-view multi-modal sensor data (camera and LiDAR) and generate control signals according to the navigation instructions based on an LLM decoder.
 
\textbf{Prediction.} VLLMs can also be utilized to predict the trajectories and actions of vehicles and pedestrians to assist in navigation. In NuPrompt \cite{wu2023language}, Wu \textit{et al.} formulate a new task by prompting to predict the described object trajectory across frames. They also propose a method PromptTrack to predict the 3D bounding box of the prompt-referred objects. BEV-InMLLM \cite{ding2024holistic} integrates BEV features and multi-view video into the LLM for perception, prediction and planning tasks. Such a method can forecast the actions of the surrounding entities and predict the potential dangers.

\subsubsection{Embodied AI}\label{embodied}
Embodied AI refers to the artificial intelligence systems designed to control physical embodiments and interact with the environments \cite{ma2024survey}. These systems possess the abilities in cognition, decision-making, and control \cite{firoozi2023foundation,hu2023toward}, which are commonly applied in the field of robotics. Since conventional robot agents mainly concentrate on some constricted tasks and lack common sense knowledge, there exists a growing number of works based on LLMs for learning versatile downstream task policies. Among these methods, VLLM-based agents constitute a significant portion and can be categorized into four streams \cite{kawaharazuka2024real}, \textit{i.e.}, perception, manipulation, planning, and navigation.

\textbf{Perception.} VLLM-based robots can extract semantic knowledge and understand environments from visual signals such as RGB or LiDAR. OpenEQA \cite{majumdar2024openeqa} introduces a new benchmark on embodied question answering perception tasks, such as object recognition, attribute recognition, object localization, \textit{etc}. This benchmark presents two types of problems: the episodic-memory task for understanding the environment through the episodic memory, and the active one by requiring only navigation actions. AffordanceLLM, 3DVG, 3D-LLM, and PaLM-E utilize 3D data to enhance perception. In AffordanceLLM~\cite{qian2024affordancellm}, the authors propose detecting the interaction point of an object (affordance grounding) via VLLM, introducing depth information to better capture object geometry and improve grounding performance. 3DVG \cite{yuan2024visual} enhances object detection and classification by introducing a language-object correlation (LOC) module that fuses 3D point cloud geometry with 2D image details via VLLM. This multi-modal fusion improves fine-grained object perception and extends the model's capacity for open-vocabulary scenarios. 3D-LLM \cite{hong20233d} proposes a framework that integrates 3D spatial information into an LLM, enabling it to process 3D point clouds and perform tasks such as 3D question answering, captioning, and navigation. Similarly, PaLM-E \cite{driess2023palm} improves embodied perception by integrating visual inputs, such as images and 3D scene representations, into the language model's embedding space using ViT and object-centric encoders. Other works utilize additional modules or agents for perception. The paper \cite{wang2024solving} introduces a multi-agent VLLM framework with specialized agents to reduce errors in object identification and coordinate refinement. REPLAN \cite{skreta2024replan} employs a VLM Perceiver, enabling robots to ground actions in visual data, allowing object detection and obstacle identification.

\textbf{Manipulation.} To create a universal robot capable of handling various downstream tasks, a key skill is the ability to manipulate objects in its environment based on the specific requirements of each task \cite{li2024foundation}. This ability can be greatly enhanced by using VLLMs. PaLM-E \cite{driess2023palm} incorporates continuous inputs like images, states and language from an agent into a pretrained LLM (PaLM \cite{chowdhery2023palm}) for manipulation planning. RT-X \cite{open_x_embodiment_rt_x_2023} introduces a large and diverse manipulation dataset ``Open X-Embodiment'' which includes 527 skills and 1M+ real robot trajectories enabling generalized policy learning. Instruct2Act \cite{huang2023instruct2act} proposes a training-free method by calling the APIs of existing foundation models using the LLM. Based on the expertise of foundation models and the reasoning ability of LLM, InstructAct can understand complex instructions for manipulation tasks. Roboflamingo \cite{li2024roboflamingo} is based on an off-the-shelf VLLM, which learns the sequential historical information with a policy head and is finetuned by imitation learning on manipulation datasets. Such a decomposition design makes Roboflamingo flexible and efficient when deploying in the real world. VoxPoser \cite{huang2023voxposer} proposes synthesizing robot trajectories for manipulation tasks based on open-set instructions and objects. Specifically, a training-free method is proposed by utilizing the code-writing abilities of LLM, which can generate 3D value maps by calling the vision-language model. Niu \textit{et al.} introduces an instruction-tuning-based method, LLARVA \cite{niu2024llarva}, which leverages structured prompts for various robotic learning tasks. They also demonstrate that visual traces, formed by intermediate 2D representations, are beneficial for aligning vision and action spaces. ManipLLM \cite{li2024manipllm} targets at predicting the contact point of the object given the text prompt, RGB image and depth map by tuning a VLLM. Kim \textit{et. al.} presents an efficient tuning VLLM OpenVLA \cite{kim2024openvla} by training on a large-scale real-world manipulation dataset.

\textbf{Planning.} In embodied AI, task planning involves decomposing high-level objectives into atomic subtasks while accounting for real-world dynamics. NLMap \cite{chen2023open} enables agents to construct open-vocabulary queryable scene representations by integrating VLMs and LLMs. It allows robots to identify relevant objects within the environment, and generate context-aware plans, thereby enhancing their ability to execute complex tasks in real-world settings. ELLM \cite{du2023guiding} assists agents in formulating and pursuing intermediate objectives,  facilitating effective exploration of the environment. By prompting LLMs with descriptions of the agent's current state, ELLM generates contextually relevant goals, guiding agents toward diverse behaviors aligned with common-sense knowledge. LLaRP \cite{szot2023large} integrates text instructions and egocentric observations to directly output actions in a dynamic environment. This approach enhances the agent's ability to execute complex tasks by learning a reinforcement learning policy. LL3DA~\cite{chen2024ll3da} directly processes 3D point cloud data by leveraging textual and visual prompts from human interactions. Equipped with visual prompts, LL3DA can enhance the agent's ability to understand, reason and plan within complex 3D environments, facilitating more effective task execution and decision-making. ConceptGraphs \cite{gu2023conceptgraphs} introduces a semantically rich 3D scene graph of the environment by leveraging 2D foundation models and the priors in LLMs. This approach enhances navigation by enabling agents to inperpret complex spatial and semantic relationships within a scene, facilitating efficient decision-making during path planning and task execution. Finally, RILA \cite{yang2024rila} utilizes multi-modal models to process sensory data, and directs an LLM-based planner to actively explore the environment, dynamically evaluating and discarding inaccurate perceptual descriptions during navigation.

\textbf{Navigation.} In the field of embodied navigation, current VLLMs combine multiple modalities to improve navigation efficiency in dynamic environments. These models integrate multi-modal data, leveraging the reasoning capabilities of LLMs to process complex instructions and accurately navigate obstacles in real-world settings~\cite{deitke2023pixavoxels}. LM-Nav~\cite{shah2022lmnav} combines an LLM (GPT3), a pretrained VLM (CLIP \cite{radford2021learning}), and a visual navigation model (ViNG \cite{shah2021ving}) to handle complex, real-world navigation. This approach enables agents to interpret natural language instructions and execute long-horizon navigation tasks without fine-tuning in challenging outdoor environments. MultiPLY \cite{hong2024multiply} introduces a model enabling active interaction with 3D environment, using multisensory inputs like visual, auditory, tactile, and thermal data. By building correlations among words, actions, and perceptions, this approach can perform context-aware actions within intricate 3D settings. EMMA \cite{yang2024embodied} learns to navigate visual environments by imitating the behavior of an LLM operating in a parallel textual world. By aligning visual observations with textual descriptions, EMMA effectively translates high-level navigation strategies from textual to visual contexts, significantly enhancing its ability to plan and execute navigation tasks in complex, real-world scenarios.

\subsubsection{Automated Tool Management}\label{tool}

The goal of automated tool management is to leverage well-established models to autonomously perform complex tasks that traditionally require human intelligence and decision making. Early efforts in automated tool management predominantly utilized pure text-based models. These models excelled in interpreting and generating textual content for various applications including programming assistance, solving mathematical problems, and executing simple logical tasks. We direct readers to a survey~\cite{qu2024tool} for an exhaust review. However, as the complexity of tasks increases, in this survey, we aim to encompass inputs with not only text but also visual signals and interactions with physical environments~\cite{koh2024visualwebarena}.

The integration of VLLMs with tool management represents a significant advancement in the field of AI. To systematically understand this integration, a taxonomy is proposed that categorizes these developments based on their system architecture, mode of integration, and application areas. This taxonomy not only organizes the existing methodologies but also provides insights into their functionalities and potential applications.

\textbf{Actions as Tool APIs.} In the context of VLLMs, tools are often conceptualized and implemented as APIs. This approach allows for modular, flexible, and scalable solutions where different functionalities can be called as needed. 

Common tools managed by VLLM planners include:
\begin{itemize}
   
\item 
\textit{Pure Vision Models}: These models, typically including CNN-based models and ViT-based models, help transform input or intermediate visual signals into vectors for further processing. 

\item \textit{Visual Language Models}: Those models, including both generative models such as VQA models, image captioning models, other VLLMs, and discriminative models such as CLIP~\cite{radford2021learning}, are the core tools for understanding and responding to queries about visual content, facillitating the planner to support applications in customer service, interactive education, and more.

\item 
\textit{Web Searching APIs}: Integrating VLLMs with web search APIs (\textit{e.g.}, google search) enhances the ability to retrieve and synthesize information from vast internet resources, making the management of information retrieval tasks more efficient.

\item
\textit{Library Calls:} VLLMs can automate the usage of software tools and libraries, facilitating tasks like code generation, software testing, and even complex design tasks in engineering and graphics.
\end{itemize}


\textbf{System Architecture.}
System architecture in VLLMs planners describes how various components like language models, visual processing units, and tool management systems are structured within the AI systems. This category helps in understanding the foundational setup of VLLMs and their operational efficiency.

\begin{itemize}
    \item \textit{Integrated Systems}: These systems exemplify a tightly coupled framework where components are seamlessly integrated to perform tasks. For instance, MM-REACT~\cite{yang2023mm} integrates textual and visual inputs to enhance multimodal reasoning and action. Similarly, AVIS~\cite{hu2024avis} combines a dynamic tool management system with planning and reasoning components to handle complex visual queries effectively.
    \item \textit{Modular Systems}: Characterized by their flexibility, these systems use distinct, interchangeable modules that can be tailored for specific tasks. Mind's Eye~\cite{liu2022mind} integrates a physics simulation engine with a language model to ground linguistic processing in physical reality, while Chameleon~\cite{lu2024chameleon} employs a sequence of plug-and-play modules for complex reasoning across multiple modalities and domains.
    \item \textit{Self-Instructing Systems}: These systems are designed to adapt and learn how to use tools autonomously through methods like self-instruction~\cite{wang2022self} and feedback loops~\cite{huang2022large,ferdinan2024into,zhang2024logicode}. GPT4Tools~\cite{yang2024gpt4tools} and Confucius~\cite{gao2024confucius} are examples where models instruct themselves to enhance their tool-using capabilities, with Confucius employing a curriculum-based learning approach to manage increasingly complex tools.
    \item \textit{Trajactory Synthesize and Self-exploration}: A recently rising pipeline utilizes the strong autonomy power of large VLLMs to explore the given physical environments, such as Visual-WebAreana~\cite{koh2024visualwebarena}, Windows-WebAreana~\cite{bonatti2024windows}, OSWorld~\cite{xie2024osworld}, and $\tau-$Bench~\cite{yao2024tau}, etc. Such exploration are usually guided by search algorithms, such as MCTS~\cite{yu2024exact}, DFS \& BFS~\cite{yang2024agentoccam}, reinforcement learning~\cite{luong2024reft}, bandits~\cite{bouneffouf2024tutorial} or trajactory sampling~\cite{anonymous2024learnbyinteract}. Typical work includes OpenWebVoyager~\cite{he2024openwebvoyagerbuildingmultimodalweb}, Agent-S~\cite{agashe2024agent}, AgentStore~\cite{jia2024agentstore}, OS-ATLAS~\cite{wu2024atlas}, and ICAL~\cite{sarchvlm}. Those synthsized tracjories are either used as knowledg-based for retrival as in-context examples, \textit{i.e.}, Retrieval-Augmented Generation (RAG) for large VLLM backbones, or distilled into smaller VLLMs for faster response~\cite{anonymous2024learnbyinteract}.
\end{itemize}

\textbf{Mode of Integration.} The mode of integration deals with how tools are incorporated and utilized within VLLMs, reflecting the operational strategies and interaction levels of these models with external resources.
\begin{itemize}
\item \textit{Direct Tool Invocation}: This straightforward approach allows models to directly call tools within their outputs. ToolkenGPT~\cite{hao2024toolkengpt}, for example, embeds tools as tokens directly within the language model's output, simplifying tool usage. CRAFT~\cite{yuan2023craft} retrieves and executes specialized code snippets directly, enabling efficient task execution. TroVE~\cite{wang2024trove}, HuggingGPT~\cite{shen2024hugginggpt} and ViperGPT~\cite{suris2023vipergpt} also share similar strategies to this.
\item \textit{Indirect Tool Utilization}: In this approach, models use tools through intermediary steps or additional processing layers, allowing for more nuanced interactions and deeper contextual understanding. MLLM-Tool~\cite{wang2024tool} and CLOVA~\cite{gao2024clova} exemplify this method, with CLOVA enhancing tool effectiveness through iterative updates based on reflective learning.
\end{itemize}

%% file: contents/text2vision.tex
\subsection{Text-to-vision}\label{text2vision}

The boundary between reality and artificiality is increasingly blurred with the rise of generative applications of VLLMs. These models signify a major shift from merely interpreting data to creatively producing new visual content. This section transitions to VLLMs' innovative visual generative applications in generating images, 3D models, and videos.

\begin{figure*} 
  \centering
  \includegraphics[width=\textwidth]{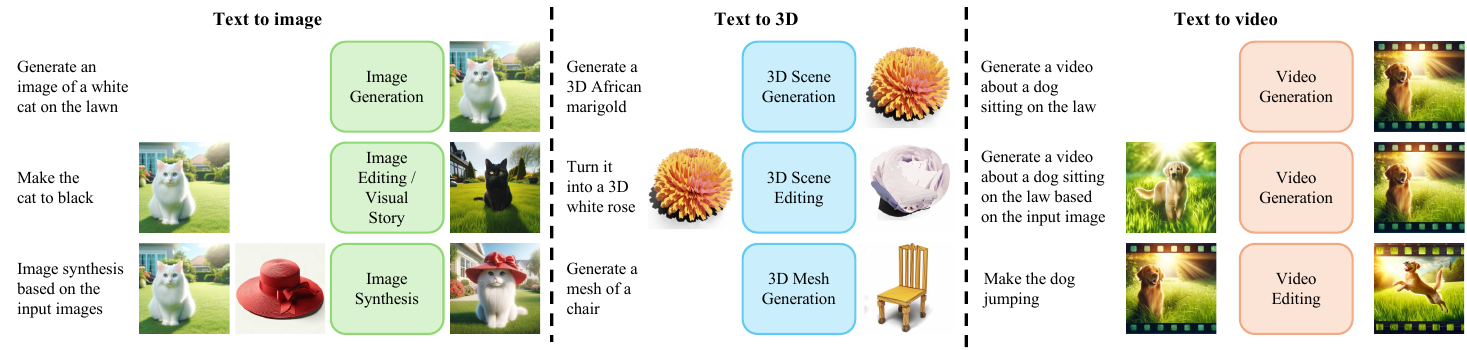}
  \caption{An illustration of text-to-vision applications.}
  \label{fig:text2vision}
\end{figure*}

\subsubsection{Text-to-image}\label{text2image}

This section focuses on how VLLMs transform text prompts into images seamlessly. Text-to-image applications include image generation, image editing, image synthesis, and visual story generation.

\textbf{Image Generation.} The core of image generation is crafting visuals from text prompts to closely reflect the prompt's intention, ensuring both visual coherence and realism. VLLMs effectively bridge the linguistic and visual realms.

Optimizing image generation prompts is crucial for synchronizing the visual and textual elements of VLLMs. GILL \cite{koh2024generating} enhances the handling of longer and more complex textual inputs for image generation by instructing LLMs to predict fixed-size visual embeddings aligned with CLIP space, thus controlling Stable Diffusion for image generation. Instead of relying on LLMs to predict fixed-size embeddings, \cite{xia2023llmga} and \cite{lian2024llmgrounded} utilize LLMs to generate more detailed prompts, resulting in more accurate and rich content.

In addition to textual refinements, some research enhances the visual embedding capabilities of VLLMs. SEED \cite{ge2023planting} and LaVIT \cite{jin2023unified} introduce sophisticated visual tokenizers that translate non-linguistic images into sequences of discrete tokens, akin to a foreign language that LLMs can interpret. These visual tokenizers enable LLMs to simultaneously see and draw. VL-GPT \cite{zhu2023vlgpt} proposes an innovative image tokenizer-detokenizer framework, which, alongside traditional text tokenization methods, allows VL-GPT to handle interleaved image-text data seamlessly.

Beyond textual or visual refinements, some research improves image generation by focusing on multimodal fusion. MiniGPT-5 \cite{zheng2023minigpt} boosts multimodal generation through "generative vokens" as pivotal elements, bridging LLMs with Stable Diffusion for more cohesive and context-aware vision-and-language interactions. Emu \cite{sun2023generative} processes multimodal inputs through an autoregressive training process. Kosmos-G \cite{pan2023kosmos} aligns the output space of MLLM with CLIP using the textual modality as an anchor. CoDi-2 \cite{tang2023codi} aligns modalities with language for both encoding and generation. \cite{aiello2023jointly} presents the Joint Autoregressive Mixture (JAM) framework to fuse existing text and image generation models into a single, robust model capable of generating seamless multimodal outputs. CM3Leon \cite{yu2023scaling} is a retrieval-augmented, token-based, decoder-only single multi-modal language model capable of generating and infilling both text and images, with notably fewer training computes compared to existing methods. 

Rather than only translating prompts into embeddings, DiffusionGPT \cite{qin2024diffusiongpt} utilizes LLMs for image generation model selection by constructing domain-specific Trees for various generative models and them to guide the selection of an appropriate image generation model.

\textbf{Image Editing.} The task of image editing involves modifying images based on language descriptions, such as object removal, color adjustment, and adding text. The primary challenge is accurately interpreting nuanced instructions while maintaining the original image's context and aesthetics.

The models in the section of Image Generation, inherently equipped for image generation, seamlessly adapt to image editing tasks when provided with image inputs. This integration leverages the rich, instruction-based datasets from InstructPix2Pix, enabling more precise and contextually accurate modifications to existing images. For example, Kosmos-G \cite{pan2023kosmos} and CoDi-2 \cite{tang2023codi} both improve image editing capability by leveraging the data constructed by InstructPix2Pix \cite{brooks2023instructpix2pix}. 

\textbf{Image Synthesis.} Image blending or synthesis involves combining elements from multiple images or generating new images from a combination of visual and textual inputs to create cohesive, novel visuals. The main challenge is achieving a seamless blend of different images while ensuring that the synthesized images are contextually relevant to the textual or visual inputs. LaVIT \cite{jin2023unified} excels in image synthesis due to its ability to autoregressively generate visual tokens from discrete tokens and reconstruct these tokens into a feature map reflecting high-level semantics. Kosmos-G's \cite{pan2023kosmos} alignment of the output space with CLIP and compositional instruction tuning enables zero-shot subject-driven image generation with interleaved multi-image and text inputs, advancing the field of image synthesis.

\textbf{Visual Stories Generation.} The task involves creating image sequences that narrate a story or follow text prompts, capturing the narrative visually. The main challenge is generating coherent and contextually relevant sequences that accurately reflect the text prompts' narrative progression. \cite{ge2023making} introduces a visual SEED tokenizer that produces discrete visual tokens carrying high-level semantics and aligns them with text, demonstrating remarkable abilities across a wide range of multimodal tasks, especially multimodal dialogue.

\subsubsection{Text-to-3D}\label{text23D}

Text-to-3D generation, driven by VLLMs, converts textual descriptions into detailed 3D models and environments, enhancing creativity and efficiency in architecture, game design, and VR. Despite challenges in spatial relationships and volumetric transformations, VLLMs excel in 3D scene generation, editing, and mesh generation.

\textbf{3D Scene Generation and Editing.} The task involves generating and modifying 3D layout to achieve desired aesthetics or functionalities, with the challenge lying in maintaining spatial coherence and realistic interactions among complex elements within dynamic scenes. 

VLLMs leverage LLMs to interpret textual instructions, provide generative feedback, conceptualize scene layouts, discretize shapes, and generate and modify interactive 3D scenes. For instance, LI3D \cite{lin2023towards} integrates LLMs as 3D layout interpreters into off-the-shelf layout-to-3D generative models, incorporating LLaVA for visual feedback to enhance the visual quality of generated content. Similarly, 3D-GPT \cite{sun20233d} decomposes complex modeling tasks into smaller components, using LLMs to interpret textual instructions into specific procedural generation tasks, enabling efficient 3D content creation without requiring direct 3D modeling by users. ShapeGPT \cite{yin2023shapegpt} adopts a "word-sentence-paragraph" framework to discretize continuous shapes into shape words, assembles these words into shape sentences, and integrates shape with instructional text for multi-modal paragraphs. GPT4Point \cite{GPT4Point} utilizes LLMs to interpret textual instructions and align them with point cloud data.

Additionally, some models incorporate additional tools or assets, such as Objaverse \cite{yang2024holodeck} or the Unity game engine \cite{de2023llmr}, to enhance their capabilities in scene understanding, task planning, self-debugging, and memory management. GPT4Point \cite{GPT4Point} introduces Pyramid-XL, an automated dataset annotation engine for generating a large-scale database of 3D object-text pairs, essential for improving its performance in understanding and generating 3D content.

\textbf{Mesh Generation.} Mesh generation translates geometric shapes into discrete vertex and edge networks for 3D modeling, facing challenges in fidelity, efficiency, and detail preservation. MeshGPT \cite{siddiqui2023meshgpt} uses a decoder-only transformer model to generate triangle meshes. By learning a vocabulary of latent quantized embeddings, MeshGPT enables the autoregressive generation of compact and detailed meshes, bypassing traditional dense methods.

\subsubsection{Text-to-video} \label{text2video}
Text-to-video generation, powered by VLLMs, transforms written narratives into engaging video sequences with high visual quality and consistency. Challenges include ensuring seamless continuity, engaging storytelling, and technical proficiency.

\textbf{Video Generation and Editing.} Video generation and editing have advanced with diffusion-based models, overcoming GAN limitations. Leveraging VLLMs, recent progress enables autonomous generation of realistic scenes and special effects, reshaping content creation.

Some researchers share a common approach of utilizing Large Language Models (LLMs) to interpret textual inputs and generate detailed scene layouts or comprehensive multi-scene scripts. LVD \cite{lian2023llm} first interprets textual inputs into detailed scene layouts, and then using these layouts to inform a diffusion model's video creation, significantly improving the fidelity and motion accuracy of generated videos. VideoDrafter \cite{long2024videodrafter} also transforms input prompts into comprehensive multi-scene scripts via LLMs, identifying common entities across scenes to maintain visual consistency through reference images generated by a text-to-image model. \cite{huang2023free} leverages LLMs for semantic prompt generation and Latent Diffusion Models (LDMs) for frame animation, aiming at producing semantically coherent, high-quality videos from textual descriptions. \cite{lin2023videodirectorgpt} utilizes the LLM to create a comprehensive video plan from a single text prompt, detailing scene descriptions, entity layouts, and backgrounds, and maintaining consistency across scenes. Mora \cite{yuan2024mora} proposes an LLM-driven multi-agent framework for generalist video generation and editing designed to mimic the capabilities of OpenAI's large-scale model, Sora. 

The other researches share a focus on enhancing the alignment between visual tokenization and the learning process of LLMs to improve video generation. MAGVITv2 \cite{yu2023language} proposes a concise and expressive image and video tokenizer, which demonstrates that LLMs can outperform diffusion models in standard video generation benchmarks. VideoPoet \cite{kondratyuk2023videopoet} synthesizes high-quality videos, specifically high fidelity motions, using a decoder-only autoregressive transformer architecture that processes multimodal inputs, including images, videos, text, and audio. SVD \cite{blattmann2023stable} also identifies and evaluates three different stages for successful training of video generation: text-to-image pretraining, video pretraining, and high-quality video fine-tuning.


%% file: tables/datasets.tex
\begin{table*}[htp]
\center
\caption{An overview of representative VLLM datasets.}  
\label{datasetList}
\begin{tabular}{lcccccccccccc}
\toprule
 \textbf{Datasets}  &\textbf{Year} &\textbf{Scale}   &\textbf{Modal} &\textbf{Task} \\ 
\hline  
SBU Captions~\cite{ordonez2011im2text}    &2011     &1M  &image-text & caption  \\  
\hline
Flickr30k~\cite{young2014image}   &2014     &145K     &image-text    &  caption	  \\
\hline
COCO~\cite{lin2014microsoft}   &2014     &567K     &image-text   & caption, vqa, rec, res \\  

\hline
Visual Genome~\cite{krishna2017visual}     &2017     &5.4M     &image-text  &  vqa, grounding  \\ 
\hline
LAION-5B~\cite{schuhmann2022laion}   &2022     &5.9B     &image-text   & caption \\  
\hline
CC3M~\cite{sharma2018conceptual}   &2018     &3.3M     &image-text   & caption	\\
\hline
CC12M~\cite{changpinyo2021conceptual}   &2021     &12.4M     &image-text   & caption	\\
\hline
COYO~\cite{kakaobrain2022coyo-700m}   &2022     &747M    &image-text   & caption	\\
\hline
Multimodal C4(MMC4)~\cite{zhu2024multimodal}   &2023     &101.2M    &image-text   & caption	\\
\hline
Obelics~\cite{laurenccon2024obelics}   &2023     &141M    &image-text   & caption \\
\hline
GRIT~\cite{peng2023kosmos}   &2023     &115M    &image-text   & caption, grounding	\\
\hline
DataComp~\cite{gadre2024datacomp}   &2023     &1.4B    &image-text   & caption	\\
\hline
MedInterp~\cite{zhou2024generalist} &2024 &13M &image-text &med-vqa, mrg \\  
\hline
MTB~\cite{moor2023medflamingo} &2024 &800K image &image-text &med-vqa \\  
\hline
PMC-vqa~\cite{zhang2023pmcvqa} &2023 &227K &image-text &med-vqa \\  
\hline
HqDC-1.4M~\cite{pang2024h2rsvlm} &2024 &1.4M &image-text &classification, vqa, grounding \\
\hline
MMShip~\cite{zhang2024popeye} &2024 &81K &image-text &grounding \\  
\hline
LHRS-Align~\cite{muhtar2024lhrsbot} &2024 &1.15M &image-text &classification, vqa, grounding \\
\hline
MMRS-1M~\cite{zhang2024earthgpt} &2024 &1M &image-text &classification, caption, vqa, grounding \\  
\hline
RSICap~\cite{hu2023rsgpt} &2023 &2.6K &image-text &caption, vqa  \\  
\hline
M-Paper~\cite{hu2024mplugpaperowl} &2024 &49K &image-text &caption, science-vqa \\  
\hline
ScienceQA~\cite{lu2022learn} &2022 &21K &image-text &science-vqa \\  
\hline
MAVIS-Caption~\cite{zhang2024mavis} &2024 &558K &image-text &caption, math-vqa \\  
\hline
MathV360K~\cite{shi2024mathllava} &2024 &360K &image-text &math-vqa \\  
\hline
GeoEval~\cite{zhang2024geoeval} &2024 &2K &image-text &math-vqa \\  
\hline
CMMaTH~\cite{li2024cmmath} &2024 &23K &image-text &math-vqa \\  
\hline
MathVerse~\cite{zhang2024mathverse} &2024 &2.6K &image-text &math-vqa \\  
\hline
WE-MATH~\cite{qiao2024wemath} &2024 &6.5K &image-text &math-vqa \\
\hline
Geo170K~\cite{gao2023gllava} &2023 &170K &image-text &math-vqa \\  
\hline
MathVista~\cite{lu2024mathvista} &2023 &6K &image-text &math-vqa \\
\hline
TextOCR~\cite{singh2021textocr} &2021 &903K &image-text &ocr \\
\hline
HierText~\cite{long2022towards} &2022 &1.2M &image-text &ocr, layout \\
\hline
TextVQA~\cite{singh2019towards} &2019 &45K &image-text &vqa \\
\hline

VIST~\cite{huang2016visual}    &2016     &72K  &image-text & Visual Storytelling \\  
\hline
VisDial~\cite{Das_2017_CVPR}   &2017     &123K     &image-text    &  Visual Dialogue, Image Grounded Dialogue\\
\hline
VizWiz~\cite{gurari2018vizwiz}   &2018     &31K     &image-text    & Caption, VQA    \\
\hline
A-OKVQA~\cite{schwenk2022okvqa}   &2022     &24K     &image-text    & VQA     \\
\hline
MSR-VTT~\cite{xu2016msr}   &2016     &10K video clips     &video-text    & video caption, video generation  \\
\hline
WebVid-10M~\cite{bain2021frozen}   &2021     &10M video clips     &video-Text    & video caption, video generation  \\
\hline
HD-VG-130M~\cite{videofactory}   &2023     &130M video clips    &video-text    & video caption, video generation   \\
\hline
ActivityNet Captions~\cite{Krishna_2017_ICCV}   &2017    &100K video clips  & video-text     & video caption, video generation   \\
\hline
HD-Vila-100M~\cite{xue2022advancing}   &2022     &103M video clips     &video-text    & video caption, video generation   \\
\hline
InternVid~\cite{wang2023internvid}   &2023     &234M video clips     &video-text    & video caption, video generation     \\
\hline
HowTo100M~\cite{miech2019howto100m}   &2019     &136M video clips     &video-text    & video caption, video generation   \\
\hline
LSMDC~\cite{rohrbach2017movie}   &2017     &128K video clips     &video-text    & video caption, video generation   \\
\hline
VATEX~\cite{wang2019vatex}   &2019     &41.3K video clips     &video-text    & video caption, video generation    \\
\hline
Open X-Embodiment~\cite{open_x_embodiment_rt_x_2023} & 2023 & 1M+ trajectories & image-action & embodied manipulation \\
\hline
OpenEQA~\cite{majumdar2024openeqa} & 2024 & 1600 questions & video-action & embodied perception \\
\hline
Talk2car~\cite{deruyttere2019talk2car} & 2019 & 850 & video-action & autonomous driving \\
\hline
nuScenes~\cite{qian2024nuscenes} & 2019 & 40K & vision (2D, 3D)-action & autonomous driving \\
\hline
HAD~\cite{kim2019grounding} & 2019 & 5.7K & video-action & autonomous driving \\
\hline
ShapeNet~\cite{chang2015shapenet}   &2015     &3M 3D models    &3D-text   & 3D generation, 3D shape retrieval  \\
\hline
Text2Shape~\cite{chen2019text2shape}   &2019     &23K 3D models   &3D-text    & 3D Generation, 3D shape retrieval  \\
\hline
Objaverse~\cite{Deitke_2023_CVPR}   &2023     &818K 3D models    &3D-text    & 3D generation, 3D shape retrieval  \\

\bottomrule

\hline
\end{tabular}
\end{table*}	

%% file: contents/challenges.tex
\section{Ethics Consideration, Challenges, and Future Work}

\subsection{Ethics Consideration}

\subsubsection{Impact on Labor Markets}

The application of VLLMs automates tasks across various industries. Sectors like creative design, customer service, and medical diagnostics may experience reduced demand for human labor, as VLLMs generate images, interpret visual data, and perform tasks traditionally requiring human expertise. This shift could displace jobs in areas such as graphic design, video editing, and medical imaging.

However, VLLMs also create new opportunities in AI oversight, ethical deployment, and AI-human collaboration. Creative professionals and medical experts may transition to supervisory roles, ensuring VLLM outputs meet brand, ethical, or medical standards. Policymakers and industries must prioritize reskilling programs to mitigate job displacement risks, promoting a balanced transition as VLLMs advance. Moreover, industries relying on manual labor face accelerated digitalization, heightening the need for responsible adoption of AI to complement, rather than replace, human workers.

\subsubsection{Bias}

VLLMs, trained on large-scale visual and textual datasets, are prone to inheriting and amplifying various biases, including sampling bias, gender/racial bias, and lingual bias. This raises ethical concerns, particularly in tasks where biased outputs can reinforce stereotypes or lead to discriminatory outcomes \cite{ruggeri-nozza-2023-multi, lee2023survey}. Addressing these biases is critical for ensuring fairness and ethical use of VLLMs across diverse domains.

\textbf{Sampling Bias.} Sampling bias in VLLMs occurs when training data is not representative of the global population. Overrepresentation of certain regions, social classes, or cultures can lead to models performing better for those groups while underperforming for others. This lack of diversity can reinforce stereotypes or inaccurately represent minorities in applications like facial recognition or content generation. 

\textbf{Gender and Racial Bias.} Gender and racial biases in VLLMs often originate from training datasets that underrepresent specific demographic groups or perpetuate harmful stereotypes. These biases manifest in biased outputs for image generation and content interpretation, as evidenced by several studies that have highlighted disparities across models and languages \cite{fraser2024examining, zhao2024gender}. Such biases are particularly concerning in high-stakes domains, including hiring, law enforcement, and healthcare, where VLLMs are increasingly relied upon for analyzing visual data. The presence of bias in these systems can result in skewed outcomes, perpetuating societal inequalities and unfair treatment. To address these issues, ongoing research is focused on developing mitigation strategies to ensure fairness in model outputs \cite{aggarwal2023fairness}.

\textbf{Lingual Bias.} Lingual bias in VLLMs arises when training data disproportionately favors dominant languages, like English, leading to better performance in those languages while underperforming in others. This bias can result in lower-quality outputs for underrepresented languages, reinforcing linguistic inequalities and limiting accessibility in applications such as multilingual visual analysis.

\subsection{Challenges}

\subsubsection{Efficiency}
The efficiency of VLLMs becomes a critical challenge due to their widespread deployment and the substantial computational demands they impose. The efficiency of VLLMs encompasses two key aspects: training and inference. 

\textbf{Training Efficiency}. For instance, training an efficient VLLM  model DeepSeek-V3 \cite{lu2024deepseek} from scratch costs approximately 5.6M dollars for 180,000 H800 GPU hours, consuming around 54,000 kWh energy and producing 25.65 metric tons of CO\textsubscript{2}. As a result, the training efficiency of VLLMs has a profound impact on economic costs, energy consumption and environment sustainability.

\textbf{Inference Efficiency}. Not only problems faced during the training phase, but inference efficiency of VLLMs is also essential, especially for real-time deployment scenarios, such as Chatbots, autonomous systems, and so on. Furthermore, inference efficiency is key to enabling deployment on edge devices with limited computational power, such as smartphones and internet-of-things devices.

\subsubsection{Interpretability and Explainability}
According to the definitions provided in the literature \cite{nauta2023anecdotal, lipton2018mythos}, \textit{interpretability} refers to the internal property, such as its parameters or structure, which does not rely on external explanations. In contrast, \textit{explainability} refers to the extrinsic ability to provide post-hoc explanations for the behavior of the models using tools or methods.

The scalability and complexity of VLLMs for addressing diverse downstream tasks present a considerable challenge in interpretability and explainability, particularly when building transparent, trustworthy, and fair models \cite{dang2024explainable}. Specifically, the intricate decision-making process, massive number of parameters, and extensive training data make VLLMs difficult to interpret and explain effectively.

\subsubsection{Hallucination}

Hallucination in VLLMs refers to the generation of outputs that are semantically plausible but factually incorrect or irrelevant to the visual input \cite{rohrbach2018object}. Hallucination in VLLMs arises due to several key factors. First, misinterpretation by vision encoders often leads to errors in recognizing spatial relationships, orientations, or subtle visual cues within images. Additionally, complex or ambiguous visual inputs—like overlapping objects, intricate backgrounds, or subtle differences in shapes and colors—can confuse the model, resulting in the generation of incorrect details. Biases in training data and insufficiently diverse examples further contribute to these errors, as the model may rely on learned patterns that do not accurately reflect the given visual context. Moreover, adversarial inputs specifically designed to exploit weaknesses in the model can exacerbate hallucinations. Lastly, inadequate training on datasets that address these challenges leaves models more prone to hallucinations \cite{huang2024visual, xu2024hallucination}. 

The dissemination of misinformation through hallucination poses significant risks, especially in sensitive applications like medical diagnostics, autonomous driving, and security systems. Erroneous outputs can lead to misinformed decisions, safety hazards, and erosion of user trust in AI systems. Furthermore, in social contexts, the propagation of misinformation can have broader societal implications, influencing public opinion and potentially causing harm. 

Addressing hallucination in VLLMs requires a multifaceted approach. Enhancing the quality and diversity of training datasets is crucial to mitigate biases and ensure balanced representation. Advanced model architectures that improve cross-modal attention mechanisms can strengthen the alignment between visual inputs and textual outputs. Techniques such as constrained decoding and incorporation of factual verification steps during generation have shown promise in reducing hallucinated content. \cite{huang2024visual,tan2024tuning}

\subsubsection{Spatial Understanding}
Recent work \cite{yang2024thinking} highlights that current VLLMs face significant challenges in spatial understanding, particularly when dealing with egocentric visual data. Spatial understanding requires that VLLMs possess the relational reasoning ability to identify spatial relationships between objects based on distance and direction. Moreover, it also requires the ability to perform egocentric-allocentric view (environment-centered) transformations. Addressing this challenge will greatly benefit the development of embodied agents and AD systems.

%% file: contents/future_direction.tex
\subsection{Future Work}
\subsubsection{Security \& Pravacy}
The increasing integration of VLLMs into diverse applications, particularly those involving sensitive visual data, necessitates a focused exploration of their security and privacy implications. Given their capabilities in image recognition, generation, and analysis, VLLMs present unique challenges and opportunities for ensuring secure and private deployments.

\textbf{Data Privacy and Confidentiality.}
VLLMs often handle large quantities of visual data that can include personal identifiers, such as faces or license plates~\cite{tu2023many,liu2024synthvlm}. Ensuring the confidentiality of such data is paramount. Future research should explore advanced data anonymization techniques tailored for images and videos, such as k-anonymity visual models~\cite{sweeney2002k,meden2018k}, where individual faces or objects are obscured but the utility for training remains intact. Additionally, leveraging encryption methods that allow VLLMs to process encrypted images without decryption (homomorphic encryption) could prevent data exposure even if security breaches occur~\cite{gupta2024enhancing,li20243d}.

\textbf{Robustness Against Adversarial Attacks.}
The visual nature of the data processed by VLLMs can make them particularly vulnerable to adversarial attacks that manipulate image pixels in a way that is intended to deceive the model~\cite{bagdasaryan2023ab,tan2024wolf}. Enhanced adversarial training that includes a wider array of manipulated visual inputs can improve model robustness~\cite{gao2024adversarial,schlarmann2023adversarial}. Development of real-time detection systems for adversarial inputs in security-critical applications, like surveillance or authentication systems, is also crucial~\cite{fares2024mirrorcheck}. These systems could utilize auxiliary networks trained specifically to detect anomalies in input images that may indicate adversarial tampering.

\textbf{Regulation and Compliance.}
Compliance with privacy regulations is critical, especially as VLLMs are deployed across geographic boundaries. Research into VLLM-specific regulatory frameworks that address the unique aspects of visual data is needed. These frameworks should guide the collection, use, and storage of visual data, as well as dictate the transparency requirements for model decisions. Future work could also explore the development of standardized privacy impact assessments for VLLMs to ensure compliance before deployment.

\textbf{Secure Model Deployment.}
Post-deployment security of VLLMs involves protecting the model from unauthorized access and ensuring that the model's operations remain secure against ongoing threats~\cite{li2024video}. Future research should investigate the use of trusted execution environments for VLLMs that can securely handle sensitive visual data on shared infrastructure. In addition, techniques such as continuous learning, in which the model updates its parameters in response to new security threats without compromising privacy, could be crucial~\cite{khan2022susceptibility,nguyen2024backdoor}.

Enhanced focus on these areas will not only fortify the security and privacy framework around VLLMs but will also facilitate their ethical and safe application in critical sectors, promoting wider acceptance and trust in these powerful technologies of users.

\subsubsection{Efficiency}
As VLLMs continue to scale, computation and deployment resources become increasingly critical. Efficiency challenges emerge during both the training and inference stages, driving more academic interest in optimizing these processes. 

\textbf{Training Efficiency}

\begin{itemize}
    \item \textbf{Token Reduction.} Visual data often contains redundant information that can be further minimized. Recent methods aim to reduce computational costs in VLLMs by pruning less important visual tokens. The objective is to preserve critical tokens while discarding or merging others with minimal performance degradation. For instance, early work such as MiniGPT-V2 \cite{chen2023minigptv2} concatenates and merges neighboring visual tokens. More recent approaches, like FastV~\cite{liu2024fastv}, leverage cross-modal guidance and attention scores to guide token reduction. LLaVolta \cite{chen2024efficient} proposes to average pool the visual tokens after a specific  LLM transformer layer, and introduces a new training paradigm that prunes different tokens at different layers during training. However, designing an automatic tuning scheme to reduce visual tokens without adding extra hyperparameters or compromising performance across different tasks remains a significant challenge.
    
    \item \textbf{Data Distillation.} VLLMs require a large amount of training data to achieve remarkable performance, as dictated by the scaling law. Consequently, the training efficiency can be significantly improved if the training data is distilled. Early research has explored methods for distilling training data for CNNs~\cite{zhao2021dataset,kim2022dataset,liu2023dream}, ViTs~\cite{zhang2023expanding, zhou2023dataset}, diffusion models~\cite{gu2024efficient,zhong2024efficient,su2024generative} and text-only LLMs~\cite{tan2024large}. However, efforts on distilling the dataset for VLLMs remain limited, making it a promising direction for further exploration.
    
    \item \textbf{Parameter-Efficient Fine-Tuning.} Parameter-efficient fine-tuning (PEFT) allows for efficient adaptation of pre-trained models to downstream tasks without fine-tuning the entire model. Techniques like Low-Rank Adaptation (LoRA)~\cite{hu2021lora} reparameterize weight matrices, reducing the number of trainable parameters and optimizing memory usage. LoRA’s variant, ReLoRA~\cite{he2022relora}, periodically updates pre-trained weight matrices with low-rank adaptors during training, further improving memory efficiency during fine-tuning. Nonetheless, current PEFT methods are mainly for LLMs, designing specific PEFT methods for VLLMs is a promising direction to explore.
\end{itemize}

\textbf{Inference Efficiency}
\begin{itemize}
    \item \textbf{Token Reduction.}
    It is also promising to reduce visual tokens during the inference phase without tuning VLLMs. LLaVA-PruMerge~\cite{zhang2024llavaprumerge} merges similar visual tokens based on the \texttt{CLS} token scores obtained from the visual encoder before the LLM decoder. However, it lacks the ability to dynamically adjust tokens based on specific instructions. Achieving effective token reduction without significant performance loss is notably more challenging than approaches that involve tuning.
    
    \item \textbf{KV Cache Optimization.}
    Key-value (KV) caching accelerates LLM inference by storing key and value states from previous tokens. However, efficient cache management for visual data is particularly critical for VLLMs based on the visual token redundancy fact. FastV~\cite{liu2024fastv} addresses this challenge by dynamically pruning unnecessary visual tokens after specific layers. Despite its effectiveness, this approach requires maintaining the KV cache for the initial predictions, resulting in memory overhead for pruned tokens. To further improve the inference efficiency of VLLMs, particularly for  tasks requiring long visual contexts, optimizing the KV cache management strategy of vision tokens during the inference phase remains a valuable area of study.
\end{itemize}
    

\subsubsection{Interpretability \& Explainability}
In order to address the challenges of interpretability, explainability and hallucinations in VLLMs, greater attention should be focused toward analyzing the interpretability and explainability of VLLMs. As highlighted in \cite{dang2024explainable}, related literature can be categorized into three perspectives, \textit{i.e.}, data, model, and training \& inference.

\textbf{Data.} Interpretability and explainability in VLLMs involve understanding how input data attributes to the models' decision-making process, and analyzing the output data of the models. Early research primarily focuses on analyzing the relationship between input visual data and class labels by visualizing the specific region of interest \cite{zintgraf2017visualizing} or attention maps \cite{park2018multimodal}. Additionally, causal inference methods \cite{morioka2023connectivity, walker2023causal} have proven to be powerful tools for exploring the connections between visual and language modalities. The advent of diffusion models has further inspired new approaches for interpretability, including generating attribution maps \cite{tang2022daam} and leveraging information-theoretic metrics \cite{liang2022high,konginterpretable}. 

\textbf{Model.} The architecture of VLLMs, including tokens, embeddings, neurons, and layers, provides valuable insights into understanding these models. Token-based methods can be classified based on token types into visual tokens \cite{neo2024towards,zhang2024redundancy} and visual-textual ones \cite{gandelsmaninterpreting,zhao2025first,li2024unified,wei2024dopra,chen2024interpreting}. These methods play a crucial role in reducing visual redundancy and mitigating hallucination issues. In contrast to specific tokens, embedding-based methods focus on coarse-grained analyses of multimodal embeddings in VLLMs, which can also be divided into visual embeddings and visual-textual embeddings. Visual embeddings \cite{verma2024cross,shi2024eagle} are good representations for understanding how VLLMs process visual information. Visual-textual embeddings \cite{dominici2023sharcs,wangfreebind,parekh2024concept} provide insights into the semantic space shared by visual data and human instructions. Neurons, the fundamental computing units in VLLMs, play a key role in understanding specific concepts or domains, which have two types: individual neurons and grouped neurons. Individual neurons \cite{dravid2023rosetta,dravid2023rosetta,liu2023cones} are associated with distinct concepts or functions. while the grouped ones \cite{schwettmann2023multimodal,kojima2024multilingual} are connected to specific tasks. Additionally, layer-based analyses focus on examining the layer architectures, such as attention heads or MLP layers from components \cite{geva2021transformer,michel2019sixteen,voita2019analyzing} or workflow \cite{kowal2024visual,palit2023towards} perspectives.

\textbf{Training \& Inference.} The training strategies and inference methods are also critical aspects for understanding VLLMs. During the training process, analyzing pretraining \cite{salin2022vision,zhou2024lima} is fundamental to understanding the subsequent instruction alignment stage.  Following pretraining, interpreting the alignment stage \cite{sun2023aligning,yan2025vigor,yu2024rlhf} is a key focus for mitigating hallucination issues and enhancing the reliability of VLLMs. Additionally, approaches such as gradient analysis \cite{mallick2024ifi,smilkov2017smoothgrad} and incorporating regularization terms \cite{dai2022plausible,huang2024opera} have shown promise in reducing hallucination during training. In the inference stage, promising directions include leveraging and explaining techniques such as Chain-of-Thought (CoT) reasoning \cite{zhang2023multimodal,ge2023chain,shao2024visual,mondal2024kam} and In-Context Learning (ICL) \cite{bansal2023rethinking,miyanishi2024multimodal}, which provide insights into the reasoning processes and decision-making capabilities of VLLMs.

\subsubsection{Complex Reasoning}
Although VLLMs have demonstrated reasoning capabilities, most current open-sourced VLLMs struggle with complex, systematic, and structured reasoning, often leading to hallucination or spatial understanding issues. To enhance their complex reasoning abilities, future research can explore techniques such as Chain-of-Thought (CoT) \cite{wei2022chain}, Tree-of-Thought (ToT) \cite{yao2024tree}, visual prompting \cite{wu2025dettoolchain}, tree search \cite{xie2024monte}, \textit{etc}. Recent advancements, such as LLaVA-o1 \cite{xu2024llava}, inspired by GPT-o1 \cite{zhong2024evaluation}, propose a multi-stage reasoning approach where each stage serves a distinct purpose in the reasoning process. To further improve inference efficiency, LLaVA-o1 incorporates stage-level beam search, generating multiple candidate results at each stage and selecting the optimal one. Investigation on using reward models to guide the planning process~\cite{luong2024reft}, or utilizing the multi-VLLM system for judgement~\cite{li2024generation} can be promising directions.

\subsubsection{Other Applications}
VLLMs can also be applied to other domains, such as face, anomaly detection, or gaming, where the integration of specialized data collection and prior domain knowledge is crucial for achieving high performance. In these tasks, VLLMs can benefit from understanding nuanced patterns and context, enabling more precise detection and classification tailored to the specific characteristics of the data.

\textbf{Face}. Privacy concerns and the scarcity of instruction-based datasets present significant challenges in acquiring high-quality, domain-specific data. As an initial effort, EmoLA~\cite{li2024facial} leverages VLLMs with a customized instruction-following dataset facial affective behavior analysis with instructions (FABA-Instruct) and tailored efficient tuning algorithms to address FABA tasks. Similarly, \cite{xing2024emo} utilizes Gemini to generate instruction data for facial expression recognition (FER) and develops a facial-prior-informed MLLM, EMO-LLaMA, to model human facial information. To further harness the language reasoning capabilities of LLMs, \cite{lan2024expllm} designs ExpLLM, incorporating accurate chain-of-thought (CoT) prompts for FER. \cite{hausladen2024social} explores the social perception of human faces with the assistance of modern VLLMs, while \cite{yang2024emollm} focuses on modeling, understanding, and responding to complex human emotions using additional video-text data and specialized MLLM models. \cite{sun2024face} introduces a practical pipeline for constructing image-text datasets with fine-grained descriptions of human faces. Building on this dataset, they propose Face-MLLM, an MLLM designed for face perception. 

\textbf{Anomaly Detection}. Data distribution bias and interpretability pose significant challenges for anomaly detection, which remain difficult for VLLMs to effectively address. In~\cite{tang2024hawk}, the authors propose Hawk as an interactive tool to interpret video anomalies precisely. \cite{zanella2024harnessing} designs a method tackling video anomaly detection (VAD) in a novel, training-free paradigm, exploiting the capabilities of pre-trained LLMs and existing vision-language models (VLMs), through VLM-based captioning to generate the text description and customized prompting to assist the reasoning. From a different research angle of data generation and evaluation, \cite{du2024uncovering} presents a comprehensive benchmark for causation understanding of video anomaly, and an evaluation metric designed to better align with human preferences.

\textbf{Gaming}. VLLMs have demonstrated significant potential in addressing gaming-related challenges by leveraging their ability to comprehend both visual and textual descriptions of gameplay. In the experiments conducted with Atari-GPT \cite{waytowich2024atari}, VLMs were employed as low-level policies in Atari games. These models exhibited advanced capabilities in processing complex visual scenes and generating strategic responses. Another application of VLMs in action role-playing games (ARPGs), such as Black Myth: Wukong, highlights their ability to operate with vision-only inputs. The proposed framework, VARP \cite{chen2024can}, successfully handled basic and medium-level combat scenarios by integrating an action-planning module with a human-guided trajectory system.

In open-world games like Minecraft, frameworks such as JARVIS-1 \cite{wang2024jarvis}, MineDreamer \cite{zhou2024minedreamer}, Adam \cite{yu2024adam}, STEVE \cite{zhao2025see}, STEVE-EYE \cite{zheng2023steve}, and MP5 \cite{qinmp5} utilize VLMs to execute task instructions, plan actions, and perform complex, long-horizon tasks. These systems combine visual and language-based planning to handle tasks ranging from resource gathering to crafting tools. Additionally, CRADLE \cite{tan2024cradle} explores the potential of VLMs in AAA games like Red Dead Redemption 2 (RDR2). CRADLE utilizes input screenshots for planning and translates these into precise keyboard and mouse controls, enabling the model to follow the storyline and successfully complete 40-minute-long missions.

\section{Conclusion}

Visual large language models (VLLMs) represent a promising frontier in the integration of vision and language, combining the strengths of both modalities into a unified framework and advancing numerous vision language applications. Despite the significant strides made in their development, the literature survey surrounding VLLMs remains limited, particularly with respect to their comprehensive application across diverse fields. In this paper, we have highlighted key application scenarios, the challenges that remain in advancing these VLLMs models, and the potential directions for future research. As VLLMs continue to evolve, addressing the existing limitations in scalability, interpretability, and cross-modal understanding will be crucial to unlocking their full potential. Ultimately, a deeper exploration of VLLMs will not only advance our ability to bridge vision and language but also open up new opportunities for multi-modal systems across a wide range of industries, from autonomous systems to healthcare and beyond. We hope this survey serves as a valuable resource for guiding future research and inspiring further innovation in the field of VLLMs.